\PassOptionsToPackage{table,dvipsnames}{xcolor}
\documentclass{muxlab}

\renewcommand*{\backref}[1]{}
\renewcommand*{\backrefalt}[4]{\ifcase#1\relax\or(page #2)\else(pages #2)\fi}

\usepackage{amsmath,amssymb,amsthm}
\usepackage{mathtools}

\crefformat{equation}{(#2#1#3)}
\Crefformat{equation}{(#2#1#3)}
\crefrangeformat{equation}{(#3#1#4)--(#5#2#6)}
\Crefrangeformat{equation}{(#3#1#4)--(#5#2#6)}
\crefmultiformat{equation}{(#2#1#3)}{ and~(#2#1#3)}{, (#2#1#3)}{, and~(#2#1#3)}
\Crefmultiformat{equation}{(#2#1#3)}{ and~(#2#1#3)}{, (#2#1#3)}{, and~(#2#1#3)}

\usepackage{url}
\usepackage{wrapfig}
\usepackage{nicefrac}
\usepackage{colortbl}

\usepackage{amsmath,amsfonts,bm}

\usepackage{algorithm}
\usepackage[noend]{algpseudocode}
\algrenewcommand\algorithmicrequire{\textbf{Input:}}
\algrenewcommand\algorithmicensure{\textbf{Output:}}
\algrenewcommand\algorithmiccomment[1]{\hfill{\color{black!55}$\triangleright$~#1}}

\newcommand{\eg}{e.g.\@}

\newcommand{\specialtoken}[1]{\text{\footnotesize\ensuremath{\langle\mathrm{#1}\rangle}}}

\def\figref#1{Fig.~\ref{#1}}

\def\secref#1{Sec.~\ref{#1}}

\def\tabref#1{Tab.~\ref{#1}}

\def\eqref#1{equation~\ref{#1}}

\def\1{\bm{1}}

\DeclareMathAlphabet{\mathsfit}{\encodingdefault}{\sfdefault}{m}{sl}
\SetMathAlphabet{\mathsfit}{bold}{\encodingdefault}{\sfdefault}{bx}{n}

\newcommand{\KL}{D_{\mathrm{KL}}}

\newcommand{\ours}{FLaRe}
\newcommand{\oursfull}{Flow-based Latent Reasoning}

\hypersetup{linkcolor=muxviolet,citecolor=muxviolet,urlcolor=muxviolet}

\graphicspath{{figures/}}

\newcommand{\mainresultsstyle}{\footnotesize\setlength{\tabcolsep}{4.3pt}}

\makeatletter
\let\LF@fs@ruled\fs@ruled
\def\fs@ruled{\LF@fs@ruled\def\@fs@pre{\normalcolor\hrule height.8pt depth0pt \kern2pt}}
\makeatother

\makeatletter
\patchcmd{\mux@maketitle}{\vskip 0.4cm}{\vskip 0.6cm}{}{\PackageError{main}{title spacing patch failed}{}}
\patchcmd{\mux@maketitle}{\vskip 0.25cm}{\vskip 0.45cm}{}{\PackageError{main}{abstract spacing patch failed}{}}
\makeatother

\title{\fontsize{17.5}{22}\selectfont What Matters for Latent Reasoning with Flow Matching}

\renewcommand{\authorlist}{%
  {\sffamily\color{muxfg}%
    \textbf{Yassine Ouali}\textsuperscript{1}\quad
    \textbf{Adrian Bulat}\textsuperscript{1,2}\quad
    \textbf{Georgios Tzimiropoulos}\textsuperscript{1,3}%
  }%
}

\renewcommand{\affiliationlist}{%
  {\small\textsuperscript{1}Samsung AI Cambridge\quad \textsuperscript{2}Technical University of Iasi\quad \textsuperscript{3}Queen Mary University of London}%
}

\abstract{Latent reasoning lets a large language model (LLM) think in a continuous space and verbalize only the answer. We argue that an effective latent thought must meet five requirements: it should be \emph{useful}, helping produce the correct answer rather than merely changing it, \emph{diverse}, so that resampling yields different reasoning trajectories, \emph{explainable}, so that a decoded chain of thought (CoT) reflects reasoning the answer actually follows, \emph{refinable} with more inference compute, and \emph{efficient}, costing less than an explicit CoT at comparable accuracy. Current methods rarely meet these requirements: they learn shortcuts from the question, distill the explicit CoT into their weights, or imitate it one token at a time. We focus on flow matching in a learned latent space, the family we argue is best placed to meet them, and identify the training choices that make it work. The result is \oursfull{} (\ours{}), a simple recipe covering what the latent space encodes and how to shape it, where to train the flow, how to read out the answer, and a final stage of training on the model's own verified thoughts. A probe for each requirement shows that \ours{} improves on prior latent methods in all five. It also compares favorably with them on arithmetic benchmarks, while reaching 97\% of the accuracy of explicit CoT at a quarter of its latency.%
}

\renewcommand{\beginappendix}{%
  \clearpage
  \appendix
  \crefalias{section}{appendix}%
  \crefalias{subsection}{appendix}%
  \section*{\huge\sffamily\bfseries Appendix}
  \addcontentsline{toc}{section}{Appendix}
  \vskip 0.5cm
}

\begin{document}

\maketitle

\section{Introduction}
\label{sec:intro}

\begin{figure}[t]
    \centering
    \includegraphics[width=\linewidth]{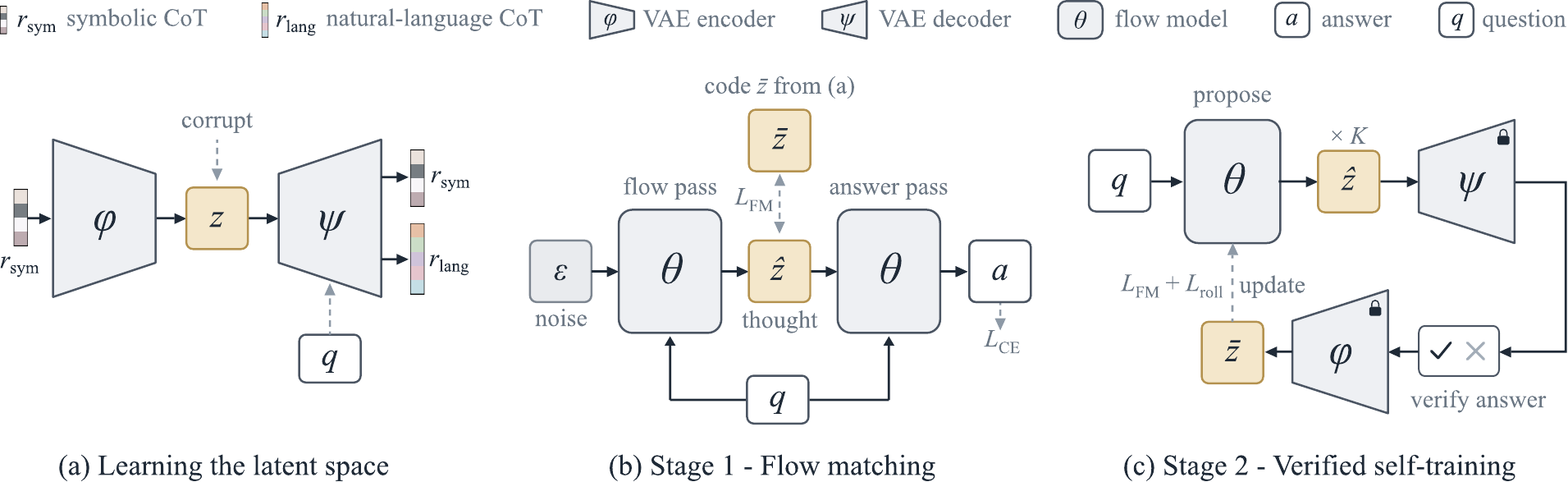}
    \caption{\oursfull{} (\ours{}). (a) A VAE encodes a symbolic CoT $r_{\mathrm{sym}}$ into a small code $\mathbf{z}$, corrupted in training for a smooth latent space, and its decoder reads it back as $r_{\mathrm{sym}}$ or, given the question $q$, as a natural language CoT $r_{\mathrm{lang}}$. (b) In stage 1, the shared flow model $\theta$ learns to denoise the code given the question and to answer from its own thought $\hat{\mathbf{z}}$. (c) In stage 2, on new questions with answers only, $\theta$ proposes $K$ thoughts per question, and those whose decoded answer is correct are re-encoded as flow targets. The model is updated with the flow loss on these targets and an answer loss $\mathcal{L}_{\mathrm{roll}}$ backpropagated through the full denoising path.}
    \label{fig:overview}
\end{figure}

Modern large language models (LLMs) increasingly rely on test-time scaling~\citep{snell2024scaling,guo2025deepseekr1,muennighoff2025s1}, reasoning at length before producing a final answer. This reasoning is usually written as an explicit chain of thought (CoT)~\citep{wei2022chain}, generated autoregressively one token at a time. While effective, explicit CoT has three limitations. Its cost grows with the length of the chain, much of this cost is spent on prose that adds little to the reasoning~\citep{hao2024coconut}, and a single rollout commits to its early tokens, so exploring another path requires further generation. Latent reasoning targets these limitations by carrying the intermediate computation in continuous states and verbalizing only the answer. In principle, sampling from a distribution of latent thoughts can explore different trajectories without generating every intermediate step in words, and thus be more efficient.

Existing latent methods differ in how they replace the token-by-token computation of explicit CoT. Horizontal methods feed continuous representations back one slot at a time~\citep{hao2024coconut,shen2025codi,tan2025colar}, while vertical methods reason within hidden states or by looping a block of layers~\citep{deng2023implicit,deng2024explicit,saunshi2025reasoning,geiping2025scaling}. Parallel methods refine a block of slots together through fixed-point iterations~\citep{wu2025pccot,kuzina2026kava} or by denoising in a learned latent space for reasoning~\citep{kang2025ladir} or text planning~\citep{lovelace2026stop}. In this work, we argue that an effective latent thought must meet five requirements: it should be useful, diverse, explainable, refinable and efficient. We focus on flow matching in a learned latent space, as in latent diffusion models~\citep{lovelace2023latent,zhang2023planner,kang2025ladir}, since it offers a direct route to four of them. Different noise draws yield different thoughts, a decoder reads each thought back as an explicit CoT, more denoising steps give a thought more compute, and a small code costs less to generate than an explicit CoT. As for usefulness, it depends on the method's design choices, which must ensure that the answer relies on the thought.

Concretely, latent flow matching starts by training a variational autoencoder (VAE) to compress CoTs into a latent space. Conditioned on the question, a flow model then refines Gaussian noise into a latent thought in this space, from which the same model generates the final answer autoregressively. In practice, with standard text diffusion, this pipeline falls far behind explicit CoT, and it works only when a few training choices are made right. We present these choices as a simple recipe, \oursfull{} (\ours{}), that follows the pipeline from the latent space to the last stage of training (\figref{fig:overview}). The VAE, trained with a dual reconstruction objective, encodes a symbolic CoT, stripped of its wording, into a small code that a fine-tuned decoder reads back both as the symbolic CoT and, given the question, as its natural language counterpart. Its latent space is shaped by corrupting the inputs and codes in training rather than by tightening reconstruction. The flow is trained mostly near pure noise, where inference starts, and with several CoTs per question. Its answer reader is trained on the flow's own predicted thoughts as well as on noised codes, so that it learns to read imperfect thoughts as generated at test time. Finally, a short stage 2 self-trains the model on its own verified rollouts for new questions, generating thoughts from pure noise as at inference and backpropagating the answer loss through the full rollout. Our main contributions are:
\begin{itemize}[leftmargin=*,itemsep=2pt,topsep=2pt,parsep=0pt]
\item \textbf{A simple training recipe.} Through controlled ablations, we identify the training choices that make latent flow matching work for reasoning, from learning the latent space to self-training on verified rollouts, and combine them into \ours{}.
\item \textbf{Requirements for latent reasoning, and probes for them.} We define five requirements an effective latent thought must meet, that it be useful, diverse, explainable, refinable and efficient, design a probe for each, and show that \ours{} improves on prior latent methods in all five.
\item \textbf{Improved accuracy and efficiency.} \ours{} compares favorably with prior latent methods on arithmetic benchmarks and reaches 97\% of explicit CoT accuracy at a quarter of its latency.
\end{itemize}

\section{Preliminaries}
\label{sec:preliminaries}

\subsection{Explicit and Latent Reasoning}
\label{subsec:explicit-latent-reasoning}

First, we provide the relevant background and key concepts of reasoning with LLMs. Given a question $q$, a model with parameters $\theta$ produces a reasoning variable $r\sim p_\theta(r\mid q)$, followed by an answer $a\sim p_\theta(a\mid q,r)$. Explicit and latent reasoning differ in the form of $r$.

\textbf{Explicit CoT.}
With explicit CoT~\citep{wei2022chain}, $r=(c_1,\dots,c_L)$ is a sequence of vocabulary tokens generated autoregressively, $p_\theta(r\mid q)=\prod_{l=1}^{L}p_\theta(c_l\mid q,c_{<l})$. We study supervised fine-tuning (SFT) on a dataset $\mathcal{D}=\{(q,r,a)\}$ of questions $q$, explicit CoTs $r$ and answers $a$. This explicit CoT training dataset will also be used to train our latent reasoning model.

\textbf{Latent reasoning.}
Here $r$ stays in a continuous space and is never verbalized during thinking. Vertical methods reason within the model's hidden states without a separate thought. Horizontal and parallel methods, including ours, instead use a latent thought $r=\mathbf{z}=(\mathbf{z}_1,\dots,\mathbf{z}_M)\in\mathbb{R}^{M\times d}$ of $M$ slots of dimension $d$, with $\mathbf{z}\sim p_\theta(\mathbf{z}\mid q)$. Horizontal methods produce the slots one at a time, while parallel methods refine all of them together. After thinking, the answer is generated token by token from the question and the latent thought, $a\sim p_\theta(a\mid q,\mathbf{z})$.

\subsection{Latent Reasoning with Flow Matching}
\label{subsec:flow-matching}

In this paper, we learn the latent thought $\mathbf{z}\in\mathbb{R}^{M\times d}$ via flow matching in a learned latent space, as in latent diffusion~\citep{lovelace2023latent,kang2025ladir}, in two steps. First, a VAE~\citep{kingma2014vae} is trained on the explicit CoT dataset to encode each CoT into a compressed latent code. Then, a flow model~\citep{lipman2023flow} is trained to generate these latent codes conditioned on the question~\citep{kang2025ladir}. At inference, the same shared flow model thinks by denoising, refining Gaussian noise into a latent thought, then produces the final answer autoregressively.

\textbf{Architecture.}
The VAE encoder, the VAE decoder and the shared flow model $\theta$, which generates both the latent thought and the answer, are all LLMs~\citep{kang2025ladir}.

\textbf{VAE training.}
Both the encoder and the decoder are initialized from a base LLM with parameters $\phi$ and $\psi$, respectively. The encoder reads $[r\;\specialtoken{Z}_1\cdots\specialtoken{Z}_M]$, an explicit CoT $r$ followed by $M$ learned slot tokens. Its final hidden states at the slot positions give the mean and log-variance of the posterior $q_\phi(\mathbf{z}\mid r)=\mathcal{N}\big(\boldsymbol{\mu}_\phi(r),\operatorname{diag}\,\boldsymbol{\sigma}^2_\phi(r)\big)$, from which the code is sampled as $\mathbf{z}=\boldsymbol{\mu}_\phi(r)+\boldsymbol{\sigma}_\phi(r)\odot\boldsymbol{\epsilon}$ with $\boldsymbol{\epsilon}\sim\mathcal{N}(\mathbf{0},\mathbf{I})$. The decoder reads $[\mathbf{z}\;\specialtoken{AE}]$, where $\specialtoken{AE}$ is a learned marker, and reconstructs $r$ from the code alone. The encoder and decoder minimize the $\beta$-VAE objective~\citep{higgins2017beta}:
\begin{equation}
  \mathcal{L}_{\mathrm{VAE}}
  =\mathbb{E}_{q_\phi(\mathbf{z}\mid r)}\big[-\log p_\psi(r\mid\mathbf{z})\big]
  +\beta\,\KL\big(q_\phi(\mathbf{z}\mid r)\,\|\,\mathcal{N}(\mathbf{0},\mathbf{I})\big),
  \label{eq:vae}
\end{equation}
where $\beta$ weights the KL term. After training, the flow targets are the posterior means, standardized per slot and dimension as $\bar{\mathbf{z}}=(\boldsymbol{\mu}_\phi(r)-\mathbf{m})/\mathbf{s}$~\citep{meshchaninov2025cosmos}.

\textbf{Flow matching.}
After VAE training, each example becomes $(q,\bar{\mathbf{z}},a)$. For $t\in[0,1]$, we interpolate from Gaussian noise $\boldsymbol{\epsilon}\sim\mathcal{N}(\mathbf{0},\mathbf{I})$ at $t=0$ to the clean VAE code $\bar{\mathbf{z}}$ at $t=1$:  $\mathbf{z}_t=(1-t)\boldsymbol{\epsilon}+t\bar{\mathbf{z}}$.
The flow model learns the constant velocity $\bar{\mathbf{z}}-\boldsymbol{\epsilon}$ along this path~\citep{lipman2023flow,liu2023rectified,li2025jit}. The shared model $\theta$ reads $[q\;\specialtoken{SOT}\;\specialtoken{t}\;\mathbf{z}_t\;\specialtoken{EOT}\;\specialtoken{SOA}\;a]$, where $\specialtoken{SOT}$ and $\specialtoken{EOT}$ bracket the thought, $\specialtoken{t}$ embeds time, and $\specialtoken{SOA}$ starts the answer~\citep{kang2025ladir}.
In the flow pass, the $M$ slots attend bidirectionally, and their final hidden states are projected to dimension $d$ to predict velocity under the flow matching loss as follows:
\begin{equation}
  \mathcal{L}_{\mathrm{FM}}
  =\mathbb{E}_{(q,\bar{\mathbf{z}}),\,t,\,\boldsymbol{\epsilon}}
  \Big[\big\|\mathbf{v}_\theta(\mathbf{z}_t,t,q)-(\bar{\mathbf{z}}-\boldsymbol{\epsilon})\big\|_2^2\Big],
  \label{eq:fm}
\end{equation}
where $t$ is drawn from a distribution over $[0,1]$ that sets which noise levels receive most of the training signal. We average the loss over four independent $(t,\boldsymbol{\epsilon})$ draws per code, as in MAR~\citep{li2024mar}, to reduce gradient variance and speed convergence. During flow training, the question is replaced by a null token $\emptyset$ with probability 0.1 to enable classifier-free guidance (CFG) at inference~\citep{ho2022classifier}. After the flow pass, the same training step runs an answer pass, which in its base form reads the clean VAE code $\bar{\mathbf{z}}$ at $t=1$ and minimizes $\mathcal{L}_{\mathrm{CE}}=-\log p_\theta(a\mid q,\bar{\mathbf{z}})$. The joint objective is $ \mathcal{L}=\lambda_{\mathrm{FM}}\mathcal{L}_{\mathrm{FM}}+\mathcal{L}_{\mathrm{CE}}$, with $\lambda_{\mathrm{FM}}$ weighting the flow loss.

\textbf{Inference.}
Given a question $q$, we start from Gaussian noise $\mathbf{z}_0=\boldsymbol{\epsilon}$ and run $S$ Euler steps from $t=0$ to $t=1$ with guidance scale $w$. The resulting latent thought $\hat{\mathbf{z}}=\mathbf{z}_1$ serves as the reasoning variable, and the same shared model generates the answer $a\sim p_\theta(a\mid q,\hat{\mathbf{z}})$ autoregressively. These steps replace the generation of an explicit CoT, while different noise draws can yield different thoughts for the same question. The frozen VAE decoder renders $\hat{\mathbf{z}}$ as an explicit CoT for offline analysis and for checking decoded final answers against references during stage 2.

\subsection{Setup}
\label{subsec:setup}

\textbf{Default setting.}
The VAE encoder and decoder are initialized from Llama-3.2-1B and Llama-3.2-3B~\citep{grattafiori2024llama3}, respectively, pretrained on OpenMathInstruct-2 (OMI-2)~\citep{toshniwal2025openmathinstruct2}, and then trained for 10 epochs on GSM8K-Aug~\citep{deng2023implicit}, an augmentation of GSM8K~\citep{cobbe2021gsm8k} with 385K questions and their symbolic CoTs (\figref{fig:paired-example}). The latent thought has $M=8$ slots of dimension $d=512$. The flow model is Llama-3.2-1B-Instruct, trained for 40 epochs on the same data with an exponential moving average (EMA) of its weights. Inference uses $S=20$ Euler steps and a guidance scale $w=4$. Appendix~\ref{app:defaults} lists all hyperparameters.

\textbf{Evaluation.}
We report GSM8K test accuracy under two readings of the same generated latent thought $\hat{\mathbf{z}}$. \textbf{Direct} accuracy uses the answer the shared flow model generates from $\hat{\mathbf{z}}$, the only reading used at inference. \textbf{Decoded} accuracy decodes $\hat{\mathbf{z}}$ into a CoT with the frozen VAE decoder and extracts its answer by string matching. The two readings measure different things. The decoded reading checks whether the model thinks correctly, that is, whether its thought solves the question, while the direct reading checks whether the model can then use this thought to answer. The decoded reading is also more forgiving, because the decoder generates the CoT token by token and can turn a latent near a valid code into a well-formed CoT, so it still relies on explicit CoT generation. The direct reading has no such correction and tests how cleanly the thought lands in the VAE space. The gap between the readings thus measures the mismatch between generated thoughts and clean VAE codes that the decoder hides. A thought that decodes to a wrong answer typically fails both readings, while one that decodes to the correct answer can still be misread and fail the direct one.

\section{Design Choices and Their Effect}
\label{sec:recipe}

\subsection{The Latent Space}
\label{subsec:latent-space}

The VAE defines the latent space used by the flow model, so we study four choices that shape this space: code capacity, corruption, CoT format and decoder routes, and decoder strength. For each VAE, we report its reconstruction on held-out CoTs, measured by exact text match, and the direct and decoded accuracies (\secref{subsec:setup}) of a flow model trained on its codes.

\begin{figure}[t]
\begin{minipage}[c]{0.48\linewidth}
    \centering
    \includegraphics[width=0.85\linewidth]{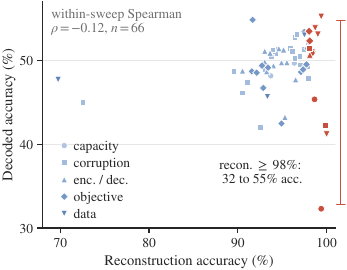}
    \caption{Reconstruction against decoded accuracy for 66 VAEs from nine sweeps.}
    \label{fig:recon-vs-acc}
\end{minipage}\hfill
\begin{minipage}[c]{0.48\linewidth}
\captionof{table}{Capacity of the latent code ($M$ slots of dimension $d$) and size and fine-tuning of the decoder $\psi$ (\%).}
\label{tab:vae-capacity}
\vspace{4pt}
\centering
\small\setlength{\tabcolsep}{3pt}
\begin{tabular*}{\linewidth}{@{\extracolsep{\fill}}lccc@{}}
\toprule
\textbf{Latent code} $M\times d$ & \textbf{Recon.} & \textbf{Direct} & \textbf{Decoded} \\
\midrule
$32\times512$ & \textbf{99.4} & 25.4 & 32.3 \\
$16\times512$ & 98.6 & 39.7 & 45.3 \\
$8\times256$ & 93.7 & 42.7 & 48.1 \\
$8\times768$ & 96.4 & 44.4 & 49.7 \\
$8\times512$ & 96.8 & \textbf{47.1} & \textbf{53.2} \\
\midrule
\textbf{Decoder $\psi$} & \textbf{Recon.} & \textbf{Direct} & \textbf{Decoded} \\
\midrule
1B, frozen & 95.3 & 37.2 & 43.4 \\
1B, fine-tuned & 93.3 & 40.8 & 46.7 \\
3B, frozen & 96.5 & 41.9 & 47.7 \\
3B, fine-tuned & \textbf{96.8} & \textbf{47.1} & \textbf{53.2} \\
\bottomrule
\end{tabular*}
\end{minipage}
\end{figure}

\begin{table}[ht]
\begin{minipage}[b]{0.48\linewidth}
\caption{Corruption of the VAE (\%).}
\label{tab:vae-corruption}
\vspace{4pt}
\centering
\small\setlength{\tabcolsep}{2pt}
\begin{tabular*}{\linewidth}{@{\extracolsep{\fill}}lccc@{}}
\toprule
\textbf{Corruption} & \textbf{Recon.} & \textbf{Direct} & \textbf{Decoded} \\
\midrule
None & \textbf{99.8} & 43.8 & 49.4 \\
No token substitution & 99.6 & 45.6 & 47.6 \\
No latent noise & 94.2 & 44.5 & 50.8 \\
No latent dropout & 96.1 & 44.8 & 51.0 \\
All three & 96.8 & \textbf{47.1} & \textbf{53.2} \\
\bottomrule
\end{tabular*}
\end{minipage}\hfill
\begin{minipage}[b]{0.48\linewidth}
\caption{Encoder input and decoder routes (\%). Reconstruction is on the input's own route, if any.}
\label{tab:vae-routes}
\vspace{4pt}
\centering
\small\setlength{\tabcolsep}{2pt}
\begin{tabular*}{\linewidth}{@{\extracolsep{\fill}}lccc@{}}
\toprule
\textbf{Encoder input, decoder routes} & \textbf{Recon.} & \textbf{Direct} & \textbf{Decoded} \\
\midrule
$r_{\mathrm{lang}}$, language route only & 43.8 & 40.9 & 44.7 \\
$r_{\mathrm{sym}}$, language route only & -- & 46.3 & 59.8 \\
$r_{\mathrm{sym}}$, symbolic route only & 98.5 & 48.6 & 55.6 \\
$r_{\mathrm{sym}}$, dual routes & \textbf{99.1} & \textbf{50.7} & \textbf{61.3} \\
\bottomrule
\end{tabular*}
\end{minipage}
\end{table}

\subsubsection{Reconstruction Alone Is Not Sufficient}
High VAE reconstruction does not by itself produce codes that the flow can generate accurately. At $d=512$, increasing the code from 8 to 32 slots raises reconstruction from 96.8\% to 99.4\% but lowers direct accuracy from 47.1\% to 25.4\% and decoded accuracy from 53.2\% to 32.3\% (\tabref{tab:vae-capacity}). Across 66 VAEs from nine sweeps that vary the capacity, corruption, objective, encoder, decoder and training data, reconstruction and decoded accuracy have a correlation of $-0.12$ (\figref{fig:recon-vs-acc}), and several of the best-reconstructing VAEs are among the worst downstream. Higher reconstruction thus makes the code more precise but harder to generate. At inference, the flow must build from the question alone a complete thought that solves the problem and remains decodable, so every detail the code preserves is one more detail to generate.

\subsubsection{Smoothness Is Necessary}
A generated thought rarely matches a clean VAE code exactly. Nearby codes should therefore decode consistently, and the decoder must tolerate approximate thoughts. Reconstruction and the KL term of \eqref{eq:vae} do not ensure this, and increasing $\beta$ degrades reconstruction before it improves generation. We thus keep $\beta=10^{-5}$ and corrupt both the encoder input and latent code during VAE training. Each input CoT token is replaced by a random token with probability $p_{\mathrm{sub}}=0.3$, while the decoder target stays clean. Half of the sampled codes receive variance-preserving noise, $\mathbf{z}\leftarrow\delta\mathbf{z}+\sqrt{1-\delta^2}\boldsymbol{\xi}$, with $\boldsymbol{\xi}\sim\mathcal{N}(\mathbf{0},\mathbf{I})$ and $\delta=0.7$. We also apply latent dropout with probability $p_{\mathrm{drop}}=0.4$~\citep{meshchaninov2025cosmos}.
All three corruptions improve direct accuracy relative to removing any one of them. Token substitution has the largest effect on decoded accuracy (\tabref{tab:vae-corruption}). Compared with no corruption, they improve both readings despite lower reconstruction. The corruptions thus do the opposite of added capacity: each CoT is encoded as a smooth region rather than a precise point, so a thought that lands nearby still decodes to it, and the flow only has to reach the region. Without them, neighboring codes decode to different CoTs, so a thought that lands near the right code keeps the plan but may be decoded to different values.

\begin{figure}[t]
\begin{minipage}[t]{0.35\linewidth}
    \centering
    \includegraphics[width=\linewidth]{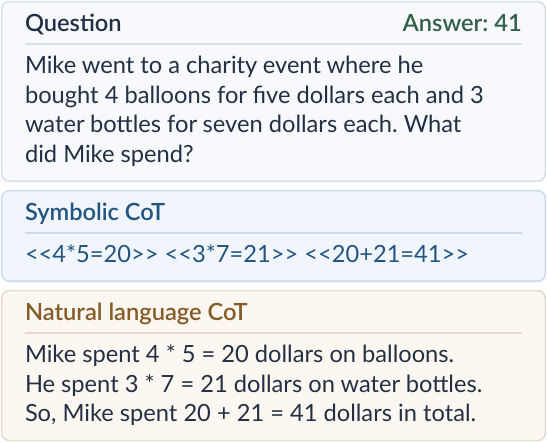}
    \caption{Symbolic (GSM8K-Aug) and aligned natural language (GSM8K-Aug-NL) CoTs of one question.}
    \label{fig:paired-example}
\end{minipage}\hfill
\begin{minipage}[t]{0.63\linewidth}
    \centering
    \includegraphics[width=\linewidth]{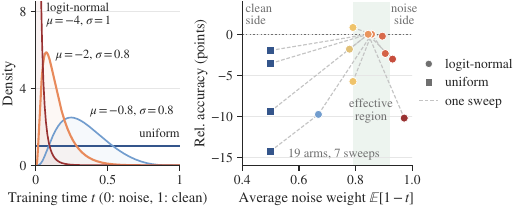}
    \caption{Time distribution of the flow loss. Left, the tested densities of $t$. Right, accuracy relative to the default of each sweep (gray lines) against the noise weight $\mathbb{E}[1-t]$, effective region in green.}
    \label{fig:time-vs-acc}
\end{minipage}
\end{figure}

\subsubsection{The Target Must Be Information Dense}
To see which style of CoT better fits latent reasoning with flow matching, we pair each question in our training data with two CoTs: a symbolic CoT $r_{\mathrm{sym}}$ that only records the reasoning steps and intermediate results needed to reach the answer, and its paired natural language CoT $r_{\mathrm{lang}}$ that explains the same steps in plain language (\figref{fig:paired-example}). The training set is thus $\mathcal{D}_{\mathrm{vae}}=\{(q, r_{\mathrm{sym}}, r_{\mathrm{lang}}, a)\}$.

The encoder then reads either CoT, and the decoder writes a CoT back through one of two routes. The symbolic route writes $r_{\mathrm{sym}}$ from the code alone, $[\mathbf{z}\;\specialtoken{AE}]$, so the code must carry every reasoning step. The language route writes $r_{\mathrm{lang}}$, and with symbolic input it also reads the question, $[\mathbf{z}\;\specialtoken{Q}\;q\;\specialtoken{AE}]$ with a learned marker $\specialtoken{Q}$, which supplies the entities, units and wording that the symbolic CoT omits. Training both routes, which we call the dual routes, gives the following dual reconstruction objective:
\begin{equation}
\mathcal{L}_{\mathrm{VAE}}
=\mathbb{E}_{q_\phi(\mathbf{z}\mid r_{\mathrm{sym}})}\big[-\log p_\psi(r_{\mathrm{sym}}\mid\mathbf{z})-\log p_\psi(r_{\mathrm{lang}}\mid\mathbf{z},q)\big]
+\beta\,\KL\big(q_\phi(\mathbf{z}\mid r_{\mathrm{sym}})\,\|\,\mathcal{N}(\mathbf{0},\mathbf{I})\big).
\label{eq:vae-dual}
\end{equation}
A VAE with a single route keeps only its own reconstruction term. In \tabref{tab:vae-routes} we compare four VAEs with different inputs and decoder routes. The results show that the symbolic CoT is the better input and a VAE that encodes the natural language CoT loses about 10 direct and 17 decoded points against the default. The symbolic CoT is also far more compressible, since the same eight slots reconstruct nearly every held-out symbolic CoT exactly but fewer than half of the natural language ones. Training the decoder on the dual routes also helps in two ways. First, the second reconstruction regularizes the code for more effective flow matching, and second, it yields a single frozen decoder that renders a latent thought as either a symbolic or a natural language CoT, which we use to probe the model. We therefore take the symbolic CoT as input and the dual routes as our default.

For datasets with only natural language CoTs, we propose to convert each CoT with an LLM that, given the symbolic language and its rules, segments the CoT into elementary steps and compiles each step into its symbolic form (Appendix~\ref{app:converter}).

\subsubsection{A Stronger Decoder Improves Downstream Accuracy}
With the 1B encoder fixed, a fine-tuned 3B decoder gives the strongest result among the decoder sizes and training choices in \tabref{tab:vae-capacity}. One possible explanation is that it lets the encoder omit details the decoder can recover, giving the flow an easier target. It may also read approximate generated thoughts more robustly.

\begin{keybox}[before skip=16pt, after skip=16pt]
\textbf{The Final Recipe.}
We carry forward an $8\times512$ VAE with symbolic CoTs as encoder input, dual decoder routes, all three corruptions, a 1B encoder, and a fine-tuned 3B decoder. \tabref{tab:defaults} lists its hyperparameters.
\end{keybox}

\subsection{The Flow Matching}
\label{subsec:flow-training}

The flow model is trained on $\mathcal{D}_{\mathrm{flow}}=\{(q,\bar{\mathbf{z}},a)\}$, where cached VAE codes replace the explicit CoTs of $\mathcal{D}_{\mathrm{vae}}$. Its training objective is $\mathcal{L}=\lambda_{\mathrm{FM}}\mathcal{L}_{\mathrm{FM}}+\mathcal{L}_{\mathrm{CE}}$, as defined in \secref{subsec:flow-matching}, with $\lambda_{\mathrm{FM}}=5$. We study four choices: where along the path to train, what the answer pass reads, which training data to use, and whether to train end to end through the full rollout.

\subsubsection{The Flow Is Trained Mostly Near Noise}
The flow loss of \eqref{eq:fm} regresses velocity at a sampled time $t$, whose distribution determines which parts of the path receive most of the training. Our sweeps vary only this distribution, comparing uniform time with logit-normals~\citep{esser2024sd3} shifted increasingly toward noise (\figref{fig:time-vs-acc}). Uniform time performs worst, and the model learns well only when training is strongly shifted toward noise, where inference starts. Otherwise, it mainly learns to refine nearly intact codes. The required shift grows with latent dimension~\citep{esser2024sd3,guo2026coladlm}, since larger codes retain more information at the same noise level. We therefore use a logit-normal time distribution, $t=\operatorname{sigmoid}(s)$ with $s\sim\mathcal{N}(-2,0.8^2)$.

\begin{table}[ht]
\begin{minipage}[b]{0.48\linewidth}
\caption{Input of the answer pass in training (\%).}
\label{tab:flow-reader}
\vspace{4pt}
\centering
\small\setlength{\tabcolsep}{3pt}
\begin{tabular*}{\linewidth}{@{\extracolsep{\fill}}lcc@{}}
\toprule
\textbf{Answer pass reads} & \textbf{Direct} & \textbf{Decoded} \\
\midrule
Exact code & 45.7 & 57.2 \\
Noised code & 46.1 & \textbf{58.0} \\
50/50 mixture, with gradients & 50.2 & 56.6 \\
Predicted only & 51.7 & 56.4 \\
50/50 mixture & \textbf{53.3} & 57.3 \\
\bottomrule
\end{tabular*}
\end{minipage}\hfill
\begin{minipage}[b]{0.48\linewidth}
\caption{Training data of the flow model, with the VAE fixed (\%).}
\label{tab:flow-data}
\vspace{4pt}
\centering
\small\setlength{\tabcolsep}{3pt}
\begin{tabular*}{\linewidth}{@{\extracolsep{\fill}}lcc@{}}
\toprule
\textbf{Flow training data} & \textbf{Direct} & \textbf{Decoded} \\
\midrule
GSM8K-Aug CoT & 51.9 & 56.7 \\
+ Diverse CoT & \textbf{54.0} & \textbf{59.7} \\
+ IID data & 51.7 & 58.5 \\
+ OOD data & 46.3 & 52.8 \\
+ IID and OOD data & 47.8 & 55.3 \\
\bottomrule
\end{tabular*}
\end{minipage}
\end{table}

\subsubsection{The Answer Pass Must Read Imperfect Thoughts}
In each training iteration, a second forward pass through the same model places a latent code $\mathbf{z}_{\mathrm{ans}}$ at the slot positions at time $t=1$ and predicts the answer tokens with the cross-entropy loss $\mathcal{L}_{\mathrm{CE}}=-\log p_\theta(a\mid q,\mathbf{z}_{\mathrm{ans}})$. \tabref{tab:flow-reader} compares five settings. Exact code reads the VAE code $\bar{\mathbf{z}}$, while noised code reads it lightly pulled toward noise. Predicted only reads the flow's one-step endpoint $\hat{\mathbf{z}}=\mathbf{z}_t+(1-t)\,\tilde{\mathbf{v}}_\theta(\mathbf{z}_t,t,q)$, computed with the guided velocity $\tilde{\mathbf{v}}_\theta$ used at inference and detached from the flow. The two mixture settings read a 50/50 split of predicted endpoints and noised codes, either detaching the endpoints or allowing the answer loss to train the flow through them.
The direct reading results in \tabref{tab:flow-reader} show that training on the model's own predictions is necessary. Since generated thoughts differ from VAE codes, a reader trained on exact or noised codes alone misreads them, and mixing predictions with noised codes 50/50 works best.
However, these predictions must stay detached, since they are noisy one-step estimates of the code, and backpropagating the answer loss through them conflicts with the flow loss.

\begin{table}[ht]
\caption{Stage 2 design (\%). Each ablation changes one part of the loop.}
\label{tab:stage2}
\vspace{4pt}
\centering
\small\setlength{\tabcolsep}{5pt}
\begin{tabular}{@{}llcc@{}}
\toprule
\textbf{Part of the loop} & \textbf{Variant} & \textbf{Direct} & \textbf{Decoded} \\
\midrule
\multicolumn{2}{@{}l}{Stage 1} & 55.3 & 61.3 \\
\multicolumn{2}{@{}l}{Stage 2} & \textbf{59.1} & \textbf{62.6} \\
\midrule
\multirow{2}{*}{Training} & No online flow loss & 59.0 & 61.6 \\
 & No stage 1 replay & 59.0 & 62.5 \\
\midrule
\multirow{4}{*}{Targets} & No verification & 57.1 & 60.7 \\
 & One trace & 58.8 & 61.8 \\
 & No harvesting & 57.2 & 61.8 \\
 & Raw latents & 58.4 & 62.2 \\
\midrule
\multirow{4}{*}{Answer loss} & No answer loss & 55.1 & 60.7 \\
 & Detached & 56.8 & 60.0 \\
 & 1-step rollout & 56.3 & 60.7 \\
 & 5-step rollout & 58.8 & 61.9 \\
\bottomrule
\end{tabular}
\end{table}

\subsubsection{Several CoTs per Question Help the Flow Model}
With the VAE fixed, we extend the flow training data, which has mostly one CoT per question, in three directions: more solutions per question written by an off-the-shelf LLM (Diverse CoT), new questions of the same style (IID data), and questions of a different style (OOD data). The new questions are the GSM8K-like and MATH-like questions of OMI-2, whose natural language CoTs we convert into symbolic ones (Appendix~\ref{app:converter}). \tabref{tab:flow-data} shows that several CoTs per question give the largest gain in both readings, while new questions help less and questions of a different style lower accuracy. With one CoT per question, different noise draws still yield different thoughts, but each question is paired with a single target code, and diffusion models trained on few samples per condition tend to memorize them~\citep{gu2025memorization,kadkhodaie2024generalization}. With several CoTs, each question has several modes, and the flow must map different noise draws to different valid solutions rather than to one code. Diverse training data is thus necessary for latent reasoning with flow matching, as it teaches the flow to cover the space of all valid solutions rather than a single one.

\subsubsection{End-to-End Training through the Full Rollout Is Required}
After stage 1, the decoded reading remains several points above the direct one. The generated thoughts therefore often contain the correct answer, but the model's readout is not adapted to them. Two limitations keep stage 1 from closing this gap. First, the answer pass trains on a mixture of noised codes and detached one-step endpoints, while inference builds thoughts from pure noise over $S$ guided steps, whose errors differ from those seen in training. Second, the answer loss updates the shared backbone but does not train thought generation through the flow steps. Addressing both requires generating thoughts from pure noise as at inference and backpropagating the answer loss through the full rollout.

In our recipe, we keep this simple and propose a stage 2 of self-training on the model's own verified rollouts, following STaR and ReST-EM~\citep{zelikman2022star,singh2024restem}. This stage uses only new questions and their reference answers, without requiring additional CoT annotations, a learned reward or a reinforcement learning objective. For each block of new questions, the EMA model generates $K=8$ thoughts per question from centered noise over $S=20$ guided steps. The frozen VAE decoder reads each thought into a symbolic CoT, and all distinct CoTs whose final answer matches the reference are retained. The retained CoTs are re-encoded into latent codes for the next round of training updates. Each update combines the flow loss on these codes and stage 1 replay with an answer loss on a fresh rollout from noise over $S=10$ guided steps. The answer loss backpropagates through every step and both guidance branches. This teaches the reader to use generated thoughts and the flow to produce thoughts that yield the reference answer. The updated EMA model then generates thoughts for the next block.

\tabref{tab:stage2} examines each part of this loop with the parent model, VAE, question pool and schedule held fixed. Stage 2 improves the direct reading by 3.8 points and the decoded reading by 1.3. These gains come from the answer loss through the full rollout, since removing it, detaching the rollout endpoint or reducing the rollout to a single step loses most or all of them. Verification is also needed, while the other choices matter less. Appendix~\ref{app:stage2} details the implementation and discusses each ablation.

\begin{keybox}[before skip=16pt, after skip=16pt]
\textbf{The Final Recipe.}
We carry forward a flow model trained on GSM8K-Aug and Diverse CoT (Appendix~\ref{app:data}), using a logit-normal time distribution shifted toward noise. In stage 1, the answer pass reads a 50/50 mixture of noised codes and the model's own detached endpoints. A short stage 2 then self-trains on verified thoughts for about one pass over a pool of new questions. Appendix~\ref{app:training} details both stages and lists all hyperparameters (\tabref{tab:defaults}), and Appendix~\ref{app:algorithms} gives their algorithms.
\end{keybox}

\section{Analysis: The Requirements of Latent Reasoning}
\label{sec:analyses}
\label{sec:properties}

Effective latent reasoning requires thoughts that are \textbf{useful}, helping produce the correct answer, and \textbf{diverse}, so that resampling can explore different reasoning trajectories for the same question. These trajectories should also be \textbf{explainable}, with recovered CoTs that reflect the reasoning the answer actually follows. At inference, the thoughts should remain \textbf{refinable} with more compute and \textbf{efficient}, costing less than explicit CoT at comparable accuracy. With our final recipe fixed, we examine these requirements through targeted probes, comparing \ours{} with Coconut, CODI and PCCoT, with explicit CoT as a reference. \tabref{tab:properties} summarizes the results, with details in Appendix~\ref{app:properties}.

\begin{table}[t]
\caption{The five requirements on GSM8K. Useful: twin pairs (\%) whose answer follows the injected thought. Diverse: gain over greedy with 16 samples. Explainable: questions (\%) whose decoded thought has all reference intermediate results in order. Refinable: gain from 0.1 to $0.5\times$ explicit CoT compute. Efficient: speedup over explicit CoT at the share of its accuracy. \ours: decoded reading, except refinable (direct / decoded) and efficient (direct, $S=2$).}
\label{tab:properties}
\vspace{4pt}
\centering
\small\setlength{\tabcolsep}{2.6pt}
\begin{tabular*}{\linewidth}{@{\extracolsep{\fill}}lcccccc@{}}
\toprule
& \textbf{Useful} & \multicolumn{2}{c}{\textbf{Diverse}} & \textbf{Explainable} & \textbf{Refinable} & \textbf{Efficient} \\
\cmidrule(lr){2-2}\cmidrule(lr){3-4}\cmidrule(lr){5-5}\cmidrule(lr){6-6}\cmidrule(lr){7-7}
\textbf{Method} & follows & pass@16 & vote & values in order & $0.1\to0.5\times$ CoT compute & speedup at \% CoT acc. \\
\midrule
Explicit CoT & 97.2 & -- & -- & -- & -- & $1.0\times$ at 100\% \\
Coconut & 7.3 & $+4.6$ & $-1.5$ & 19.1 & $+2.8$ & $2.6\times$ at 61\% \\
CODI & 31.8 & $+3.5$ & $-0.7$ & 38.5 & $+6.1$ & $2.0\times$ at 94\% \\
PCCoT & 27.7 & $+0.8$ & $-1.9$ & 7.8 & $-10.5$ & $3.9\times$ at 91\% \\
\midrule
\ours, stage 1 & 39.7 & $+16.6$ & $+3.3$ & 46.0 & $+0.4$ / $+18.2$ & $3.9\times$ at 93\% \\
\ours, stage 2 & 37.3 & $+14.7$ & $+1.8$ & 48.6 & $+1.8$ / $+24.2$ & $3.9\times$ at 97\% \\
\bottomrule
\end{tabular*}
\end{table}

\subsection{Useful Latent Thoughts}
A latent thought should carry question-specific computation that helps produce the correct answer. For Coconut and CODI, prior work reports little change in accuracy when thoughts are removed, especially on logical reasoning tasks~\citep{rizvimartel2026illusion,dilgren2026interpretable}, or, for Coconut, swapped across questions~\citep{zhang2025dolatent}. Training by curriculum or distillation can internalize the CoT in the model's weights, allowing a shortcut from the question to the answer~\citep{zhang2025dolatent,cui2026weak,aswal2026observable}. We probe this dependence with a \emph{twin test}: changing one number in a question produces a twin with a different answer. We inject the model's thought for the twin into the original question and measure how often the answer matches the twin's answer. Explicit CoT follows the twin's CoT on 97\% of pairs, compared with 40\% and 37\% for our two stages, 32\% for CODI, 28\% for PCCoT and 7\% for Coconut (\tabref{tab:properties}). The harder twins often produce incorrect thoughts. Among pairs that each model answers correctly on both questions before injection, our following rates rise to 91\% in stage 1 and 76\% in stage 2. Zeroing the thought or giving the reader only the question also reduces accuracy in both stages, supporting the thoughts' contribution to correct answers (Appendix~\ref{app:useful}).

\subsection{Diverse Latent Thoughts}
Resampling should explore different valid trajectories for the same question, providing coverage that voting, verification, reinforcement learning or self-training can turn into higher accuracy. Coconut, CODI and PCCoT generate a single thought deterministically, and the superposition Coconut was designed for collapses after fine-tuning~\citep{rizvimartel2026illusion}. New candidates therefore require perturbing the thought at inference~\citep{wang2025inference,wang2026gts}. We probe this diversity with a \emph{sampling test} of 16 thoughts per question, using independent noise draws for our flow and noise perturbations for the baselines. We measure coverage as pass@16, the share of questions solved by at least one sample, and test whether majority voting improves on the greedy answer. The baselines gain at most 5 points of coverage, and voting never improves on greedy. On the decoded reading, our stage 1 model gains 16.6 points of coverage and 3.3 through voting, while stage 2 gains 14.7 and 1.8, respectively. Stage 2 retains most of this coverage while improving single-sample accuracy (Appendix~\ref{app:diverse}).

\subsection{Explainable Latent Thoughts}
An explainable latent thought should decode into an explicit CoT that reflects the reasoning the answer follows. The baselines lack a dedicated thought decoder, so their thoughts are read through the LM head, since a thought fed back as an embedding often behaves like its nearest token~\citep{rizvimartel2026illusion}. This recovers only fragments~\citep{shen2025codi}, and full trajectories require an explicit search over candidate traces~\citep{dilgren2026interpretable}, which still does not show that the answer follows them~\citep{aswal2026observable}. We probe recovery with a \emph{reading test} that checks whether every reference intermediate result appears in order, in the top-1 LM-head tokens for the baselines and as complete values in the CoT decoded by the frozen VAE decoder for ours. Our thoughts meet this criterion on about half of the questions, more often than any baseline (\tabref{tab:properties}). We then probe faithfulness on twin pairs whose edit changes an intermediate value. Our decoded thoughts usually recover the twin's value, and the answer follows this recovered reasoning far more often than for the baselines (Appendix~\ref{app:explainable}).

\subsection{Refinable Latent Thoughts}
More inference compute should refine a latent thought until accuracy saturates, as in test-time scaling of explicit CoT~\citep{snell2024scaling,muennighoff2025s1}. Existing methods are constrained by their training budgets: Coconut becomes unstable beyond two thoughts per reasoning step~\citep{hao2024coconut}, CODI peaks at six latent tokens~\citep{shen2025codi}, and PCCoT at three Jacobi iterations~\citep{wu2025pccot}. We probe this with a \emph{budget sweep} of each frozen model, including budgets beyond those used in training. In \tabref{tab:properties}, raising the budget from a tenth to half that of explicit CoT adds at most 6.1 points to the baselines, and PCCoT even loses accuracy, while our decoded reading gains 18.2 and 24.2 points in stages 1 and 2 and the direct reading gains at most 1.8. Since the flow is trained across noise levels, more Euler steps keep refining its thought, while the baselines peak at or below their training budgets (Appendix~\ref{app:refinable}).

\subsection{Efficient Latent Thoughts}
Since latent reasoning does not verbalize every step, it should cost less than explicit CoT at comparable accuracy. We probe this trade-off with a \emph{latency test} of accuracy and time to answer on one GPU, one question at a time. In \tabref{tab:properties}, the best baselines reach 94\% of explicit CoT accuracy at a 2.0$\times$ speedup (CODI) or 91\% at 3.9$\times$ (PCCoT), while our stage 2 model, with the direct reading at two Euler steps, reaches 97\% at 3.9$\times$ (Appendix~\ref{app:efficient}).

\section{Comparison with Prior Latent Reasoning Methods}
\label{sec:comparison}

\begin{table}[t]
\caption{Direct answer accuracy (\%) of models trained on GSM8K-Aug, on GSM8K (IID) and on GSM8K-Hard, SVAMP and MultiArith (OOD). Bold: best latent method per column and scale. Baselines come from the original papers when reported, else from our runs (Appendix~\ref{app:baselines}).}
\label{tab:main}
\vspace{4pt}
\centering
\providecommand{\mainresultsstyle}{\scriptsize\setlength{\tabcolsep}{4.5pt}}%
\mainresultsstyle
\begin{tabular}{@{}lcccc|cccc|cccc@{}}
\toprule
& \multicolumn{4}{c}{\textbf{Qwen2.5-0.5B-Instruct}} & \multicolumn{4}{c}{\textbf{Llama-3.2-1B-Instruct}} & \multicolumn{4}{c}{\textbf{Llama-3.2-3B-Instruct}} \\
\cmidrule(lr){2-5}\cmidrule(lr){6-9}\cmidrule(lr){10-13}
& IID & \multicolumn{3}{c}{OOD} & IID & \multicolumn{3}{c}{OOD} & IID & \multicolumn{3}{c}{OOD} \\
\cmidrule(lr){2-2}\cmidrule(lr){3-5}\cmidrule(lr){6-6}\cmidrule(lr){7-9}\cmidrule(lr){10-10}\cmidrule(lr){11-13}
\textbf{Method} & GSM8K & Hard & SVAMP & MArith & GSM8K & Hard & SVAMP & MArith & GSM8K & Hard & SVAMP & MArith \\
\midrule
\rowcolor{gray!8} Explicit CoT & 59.7 & 17.0 & 61.6 & 98.3 & 62.5 & 14.7 & 66.5 & 97.8 & 72.4 & 20.8 & 73.4 & 98.3 \\
\rowcolor{gray!8} No-CoT & 27.3 & 6.4 & 38.5 & 57.8 & 35.0 & 7.4 & 38.1 & 67.8 & 39.2 & 9.3 & 56.7 & 96.7 \\
\midrule
Coconut & 28.2 & 6.8 & 41.9 & 58.3 & 36.1 & 8.3 & 48.4 & 87.2 & 45.3 & 10.9 & 57.5 & 96.7 \\
iCoT & 32.9 & 7.4 & 41.0 & 66.7 & 36.5 & 8.4 & 40.8 & 72.8 & 42.1 & 9.6 & 50.7 & 91.1 \\
LaDiR & 39.3 & 9.0 & 44.5 & 88.3 & 40.6 & 8.9 & 38.6 & 81.7 & 41.5 & 10.2 & 49.4 & 94.4 \\
PCCoT & 30.6 & 7.3 & 36.3 & 67.2 & 53.4 & 12.8 & 57.5 & 97.2 & 57.4 & 13.8 & 70.9 & \textbf{100.0} \\
CODI & 39.1 & 9.1 & 46.9 & \textbf{90.0} & 55.6 & 12.8 & \textbf{61.1} & 96.1 & 59.5 & 14.9 & 71.2 & 99.4 \\
KaVa & 46.9 & 10.8 & 50.6 & -- & 56.5 & 12.7 & 58.9 & -- & \textbf{65.7} & \textbf{15.2} & \textbf{72.7} & -- \\
\midrule
\ours, stage 1 & 50.0 & 11.7 & 49.1 & 78.3 & 55.3 & 12.5 & 57.0 & 94.4 & 62.5 & 14.8 & 63.8 & 98.3 \\
\ours, stage 2 & \textbf{53.8} & \textbf{13.0} & \textbf{51.0} & 88.3 & \textbf{59.1} & \textbf{13.7} & 58.8 & \textbf{98.3} & 65.2 & \textbf{15.2} & 66.2 & 99.4 \\
\bottomrule
\end{tabular}
\end{table}

Following CODI and KaVa, \tabref{tab:main} reports direct accuracy on GSM8K and zero-shot transfer to GSM8K-Hard~\citep{gao2023pal}, SVAMP~\citep{patel2021svamp} and MultiArith~\citep{roy2015multiarith}. \ours{} follows the final recipe of \secref{subsec:flow-training}, whose Diverse CoT and stage 2 questions help adapt the LLM to flow matching, a task it was not pretrained for, while all other methods except LaDiR build on its autoregressive abilities. Even stage 1, trained on the same GSM8K-Aug and Diverse CoT rows as LaDiR, our most direct competitor, which also denoises in a learned latent space, gains 10.7 to 21.0 points on GSM8K, so the gap comes from the design choices of \secref{sec:recipe} rather than from the data. Stage 2 then improves every column at every scale, by 2.7 to 3.8 points on GSM8K. It leads all latent methods on GSM8K at 0.5B and 1B, by 6.9 and 2.6 points over KaVa, performs comparably to it at 3B, and leads or ties on GSM8K-Hard at every scale. SVAMP is the main exception, likely because nearly half of its questions add distractor numbers, which are rare in GSM8K-Aug.

\section{Related Work}
\label{sec:related}

\tabref{tab:positioning} first places each method on the design axes of the paper: the family, where the thought lives, how it is produced, what supervises it, whether it can be decoded and how, and whether the inference budget is fixed at training or adjustable. Earlier methods keep the thought in the model's embeddings or hidden states, produce it one slot at a time or through a fixed number of iterations, learn it from an explicit CoT teacher, and read it, if at all, through the LM head at a budget fixed at training. The diffusion family and ours generate the thought in a learned latent space by denoising, with an adjustable budget, and decode it through the VAE decoder.

\textbf{Latent reasoning.} Latent reasoning methods~\citep{chen2025reasoning,zhu2025survey} move the intermediate computation of explicit CoT~\citep{wei2022chain} from discrete tokens into an implicit hidden computation, conducted either vertically over a fixed set of input slots, within the hidden states~\citep{deng2023implicit} or by looping~\citep{saunshi2025reasoning}, or horizontally with added tokens or blocks of tokens, without generating an intermediate text sequence at inference.

\begin{table}[t]
\caption{Latent reasoning methods along the design axes of the paper.}
\label{tab:positioning}
\vspace{4pt}
\centering
\fontsize{8.5}{10.5}\selectfont\setlength{\tabcolsep}{3pt}
\begin{tabular*}{\linewidth}{@{\extracolsep{\fill}}lllllll@{}}
\toprule
\textbf{Method} & \textbf{Family} & \textbf{Thought lives in} & \textbf{Produced by} & \textbf{Trained with} & \textbf{Decodable} & \textbf{Budget} \\
\midrule
Explicit CoT & Horizontal & Tokens & AR decoding & CoT SFT & Written out & Model-chosen \\
iCoT & Vertical & Hidden states & Single pass & CoT curriculum & No & Fixed \\
Pause tokens & Vertical & Hidden states & Single pass & Answer loss & No & Fixed \\
Coconut & Horizontal & Embeddings & AR feedback & CoT curriculum & LM head & Fixed \\
CODI & Horizontal & Embeddings & AR feedback & CoT distillation & LM head & Fixed \\
PCCoT & Parallel & Embeddings & Jacobi iteration & CoT distillation & LM head & Fixed \\
KaVa & Parallel & KV cache & Jacobi iteration & KV distillation & Not reported & Fixed \\
Looped models & Vertical & Hidden states & Looping & LM loss, no CoT & No & Adjustable \\
LaDiR & Parallel & Latent space & Denoising & Diffusion on codes & VAE decoder & Adjustable \\
\midrule
\textbf{\ours} & \textbf{Parallel} & \textbf{Latent space} & \textbf{Denoising} & \textbf{Flow matching, self-training} & \textbf{VAE decoder} & \textbf{Adjustable} \\
\bottomrule
\end{tabular*}
\end{table}

\textbf{Implicit CoT.} The first family transfers the thinking from explicit next tokens into the hidden states~\citep{deng2023implicit,deng2024explicit,yu2024distilling}, as models already perform some multi-hop reasoning latently~\citep{yang2024large,biran2024hopping,lindsey2025biology}.

\textbf{Filler tokens.} Pause or filler tokens appended between the input and the answer~\citep{goyal2024pause,pfau2024dot}, or after each word in recurrent language models~\citep{herel2024thinking}, widen the parallel computation before the answer without adding sequential feedback steps, akin to register tokens in vision Transformers~\citep{darcet2024registers}.

\textbf{Continuous feedback.} Autoregressive feedback methods, popularized by Coconut~\citep{hao2024coconut}, keep the mechanism of explicit CoT but feed the continuous representation of each step back as the next input instead of a discrete token, as in CODI~\citep{shen2025codi} among others~\citep{cheng2024compressed,zhang2025soft,tan2025colar,wei2025simcot}. These methods are analyzed theoretically by \citet{zhu2025superposition}.

\textbf{Looped computation.} Looped or recurrent methods~\citep{dehghani2019universal,giannou2023looped,fan2025looped,saunshi2025reasoning,geiping2025scaling,zhu2025ouro} keep the number of slots fixed and apply the same layers over them several times, so the reasoning becomes an iterative computation within the hidden states~\citep{wang2025hrm,jolicoeur2025trm,baek2026gram}. Newer methods loop over a fixed block of slots in parallel instead of one step at a time, as in PCCoT~\citep{wu2025pccot} and KaVa~\citep{kuzina2026kava}.

Whatever their use of computation, most of these methods still use the explicit CoT in training, through a curriculum as in Coconut and stepwise internalization~\citep{deng2024explicit}, or distillation~\citep{deng2023implicit} as in CODI and KaVa.

\textbf{Diffusion.} Latent diffusion models for text~\citep{lovelace2023latent,zhang2023planner,meshchaninov2025cosmos} generate by denoising a latent variable from Gaussian noise. In reasoning and planning, the latent is denoised, conditioned on the input, into a state that encodes a step or the full trajectory, followed by autoregressive generation, as in LaDiR~\citep{kang2025ladir} and \citet{lovelace2026stop}, with the latent space induced by a VAE trained on explicit CoT~\citep{kang2025ladir}.

\section{Conclusion}
\label{sec:conclusion}

In this paper, we studied what matters for latent reasoning with flow matching through controlled ablations of every stage of the pipeline, and distilled the answers into a simple recipe. We then set out five requirements a latent reasoning method should meet, useful, diverse, explainable, refinable and efficient, designed a probe for each, and found that the thoughts of \ours{} improve on those of prior latent methods in all five.
\ours{} also compares favorably with them on arithmetic benchmarks, while reaching 97\% of the accuracy of explicit CoT at a quarter of its latency.

\beginappendix

\titlecontents{lsection}[1.6em]{\addvspace{1.5pt}\small}{\contentslabel{1.6em}}{\hspace*{-1.6em}}{\titlerule*[0.7pc]{.}\contentspage}
\titlecontents{lsubsection}[3.9em]{\small}{\contentslabel{2.3em}}{\hspace*{-2.3em}}{\titlerule*[0.7pc]{.}\contentspage}
\startcontents[appendix]
{\setlength{\parskip}{0pt}\printcontents[appendix]{l}{1}{\setcounter{tocdepth}{2}}}
\vspace{10pt}

\section{Method and Experimental Details}
\label{app:training}

\subsection{Implementation Details}
\label{app:defaults}
\label{app:other-details}

\tabref{tab:defaults} lists the full details and hyperparameters of the final recipe, and Appendix~\ref{app:algorithms} gives its algorithms.

\begin{table}[ht]
\caption{Implementation details of the final recipe: the VAE, the flow model of stages 1 and 2, the inference and the evaluation protocol.}
\label{tab:defaults}
\vspace{4pt}
\centering
\scriptsize\setlength{\tabcolsep}{2pt}
\newcommand{\blockhead}[1]{\multicolumn{4}{@{}l}{\textbf{#1}}\\[1pt]}
\newcommand{\blocknote}[1]{\multicolumn{4}{@{}p{\linewidth}@{}}{#1}\\\addlinespace[6pt]}
\renewcommand{\specialtoken}[1]{\text{\tiny\ensuremath{\langle\mathrm{#1}\rangle}}} %
\begin{tabular}{@{}ll@{\hskip 10pt}ll@{}}
\toprule
\blockhead{VAE}
\blocknote{The VAE is initialized from a model trained for one epoch on the natural language CoTs of OMI-2~\citep{toshniwal2025openmathinstruct2} with 32 slots and a single natural language route. It is then reduced to $M=8$ slots at initialization and trained with the dual objective of \eqref{eq:vae-dual} with both terms weighted equally. Token substitution draws the replacement uniformly from the vocabulary with clean decoder targets, and the latent noise and dropout act on the sampled code before decoding. The statistics $(\mathbf{m},\mathbf{s})$ are the mean and standard deviation of the posterior means of 20K training rows.}
Backbones $\phi$, $\psi$ & Llama-3.2-1B / 3B, fine-tuned & Token substitution $p_{\mathrm{sub}}$ & 0.3 on the encoder input \\
Initialization & OMI-2 language VAE, $32\times512$ & Latent noise & VP, $\delta=0.7$, on half of the codes \\
Latent code $M\times d$ & $8\times512$ & Latent dropout $p_{\mathrm{drop}}$ & 0.4 \\
Input, decoder routes & $r_{\mathrm{sym}}$, dual routes, equal weights & Optimizer & AdamW, $(0.9,0.98)$, wd $10^{-5}$, clip 1.0 \\
Training data & paired GSM8K-Aug, 367K rows & Learning rate & $10^{-4}$ encoder, $2\cdot10^{-5}$ decoder \\
Epochs, batch, sequence & 10, 256, 512 tokens & Schedule & cosine, 5\% warmup \\
KL weight $\beta$, free bits & $10^{-5}$, 2.0 nats & Statistics $(\mathbf{m},\mathbf{s})$ & per slot and dimension, 20K rows \\
\midrule
\blockhead{Flow model, stage 1}
\blocknote{The flow and answer passes map a latent into the hidden space of $\theta$ through separate projectors, a LayerNorm followed by a two-layer MLP per slot, with a slot embedding shared by both passes. The time enters as a sinusoidal embedding at the $\specialtoken{t}$ position and through an adaptive LayerNorm in the zero-initialized velocity head. The learning rate decays with a cosine over the last fifth of training, and the EMA is a single-precision shadow updated after every step.}
Backbone $\theta$ & Llama-3.2-1B-Instruct & EMA decay & 0.9999 \\
Input sequence & $[q\;\specialtoken{SOT}\;\specialtoken{t}\;\mathbf{z}_t\;\specialtoken{EOT}\;\specialtoken{SOA}\;a]$ & Loss weights & $\lambda_{\mathrm{FM}}=5$, answer loss 1 \\
Prediction target & velocity $\bar{\mathbf{z}}-\boldsymbol{\epsilon}$ & Time distribution & logit-normal, $s\sim\mathcal{N}(-2,0.8^2)$ \\
Latent projectors & separate for the two passes & Draws per code $n$ & 4, independent $(t,\boldsymbol{\epsilon})$ \\
Slot, time embedding & shared by both passes & Question drop (CFG) & 0.1 \\
Training data & GSM8K-Aug 385K + Diverse CoT 216K rows & Answer pass input & noised code or detached endpoint \\
Training updates & 60K & $\rho_{\max}$, endpoint $w$ & 0.3, 4 \\
Batch, sequence & 256, 512 tokens & Endpoint share $p_{\mathrm{pred}}$ & 0.5 \\
Optimizer & AdamW, $(0.9,0.95)$, wd $10^{-5}$, clip 1.0 & Hardware & 16 GPUs, FSDP, bf16 \\
Learning rate & $10^{-4}$, warmup-stable-decay & Seed & 42 \\
Warmup, decay & first 1,000, last 20\% of the updates & & \\
\midrule
\blockhead{Flow model, stage 2}
\blocknote{The loop of stage 2 runs in blocks: every $B$ updates, the EMA proposer draws $K$ thoughts for each question of a new block, the decoded CoTs whose final value matches the reference are re-encoded with the frozen VAE, and the next $B$ updates minimize the flow loss on these codes, mixed with stage 1 replay, together with the answer loss backpropagated through the full rollout from noise. The proposer uses the EMA weights, and the final stage 2 model is the LIVE one.}
Question pool & 100K OMI-2 questions, answers only & Replay $\alpha$, draws $n$ & 0.25, 4 sharing one noise \\
Block $N$, updates $B$ & 1,024 questions, 16 updates & Rollout & $S_{\mathrm{train}}=10$, $w=4$, all steps \\ 
Proposals $K$, proposer & 8, EMA weights (decay 0.999) & Loss weights & $\lambda_{\mathrm{FM}}=5$, answer loss 1 \\
Proposal sampler & Euler, $S=20$, $w=4$, noise 1.0 & Optimizer & AdamW, $(0.9,0.95)$, wd $10^{-5}$, clip 1.0 \\
Verification & decoded final value $=$ reference & Learning rate & $3.5\cdot10^{-6}$, cosine, 1\% warmup \\
Kept per question & all distinct accepted, at most $K$ & Updates & 1,600 (100 blocks) \\
Targets & re-encoded posterior means & Precision & fp32 weights, bf16 autocast, DDP \\
Batch per update & 64 block rows, 16 replay rows & Final model & LIVE weights, not the EMA \\
\midrule
\blockhead{Inference}
\blocknote{The initial noise is seeded by the evaluation seed and the question index, so every model sees the same noise, and is integrated with the guided Euler update of \eqref{eq:guided-euler}. The same model then generates the answer greedily from $\hat{\mathbf{z}}$ with the question visible, and the decoded reading maps $\hat{\mathbf{z}}$ back with $(\mathbf{m},\mathbf{s})$ and decodes it greedily with the frozen decoder, without the question.}
Solver & Euler, $S=20$, uniform grid & Direct reading & greedy, at most 64 tokens \\
Guidance $w$, noise scale & 4, 1.0 & Decoded reading & frozen $\psi$, greedy, at most 512 tokens \\
Initial noise & one canonical seed per question & & \\
\midrule
\blockhead{Evaluation}
\blocknote{Accuracies are over the full test sets with one generation per question, the prediction being the last number of the answer or of the decoded CoT, correct when it equals the reference within $10^{-4}$. Stage 1 is evaluated with the EMA weights and stage 2 with the LIVE weights. The OOD evaluation sets run zero-shot with the same prompt, sampler, seed and grader as GSM8K: GSM8K-Hard~\citep{gao2023pal}, SVAMP~\citep{patel2021svamp} and MultiArith~\citep{roy2015multiarith}.}
Answer extraction & last number, match at $10^{-4}$ & Stage 1 weights & EMA \\
Test sets & GSM8K 1,319, GSM8K-Hard 1,319 & Stage 2 weights & LIVE \\
 & SVAMP 1,000, MultiArith 180 & & \\
\bottomrule
\end{tabular}
\end{table}

\subsection{Flow Model, Stage 1}

\subsubsection{Answer Pass}
\label{app:answer-path}

\textbf{Two inputs.}
Each training row contributes one answer pass in which the slot positions hold a latent $\mathbf{z}_{\mathrm{ans}}$ at time $t=1$. The row uses the model's own one-step endpoint with probability $p_{\mathrm{pred}}=0.5$ and the noised code otherwise,
\begin{equation}
\mathbf{z}_{\mathrm{ans}}=
\begin{cases}
\hat{\mathbf{z}} & \text{if } u<p_{\mathrm{pred}},\\
(1-\rho)\,\bar{\mathbf{z}}+\rho\,\boldsymbol{\epsilon} & \text{otherwise},
\end{cases}
\qquad u\sim\mathcal{U}[0,1].
\end{equation}

\textbf{Noised code.}
When the answer pass takes the cached code as input, the code is pulled toward noise with $\rho\sim\mathcal{U}[0,\rho_{\max}]$, $\rho_{\max}=0.3$ and $\boldsymbol{\epsilon}$ the first noise draw of the flow loss for that row. This is done to simulate the imperfect thoughts the model generates at inference.

\textbf{One-step endpoint.}
When the answer pass takes the model's own thought as input, $\mathbf{z}_{\mathrm{ans}}=\hat{\mathbf{z}}$ is the one-step estimate of the clean code under classifier-free guidance,
\begin{equation}
\begin{aligned}
\hat{\mathbf{z}}&=\mathbf{z}_t+(1-t)\,\tilde{\mathbf{v}}_\theta(\mathbf{z}_t,t,q),\\
\tilde{\mathbf{v}}_\theta(\mathbf{z}_t,t,q)&=\mathbf{v}_\theta(\mathbf{z}_t,t,\emptyset)+w\,\big[\mathbf{v}_\theta(\mathbf{z}_t,t,q)-\mathbf{v}_\theta(\mathbf{z}_t,t,\emptyset)\big],
\end{aligned}
\end{equation}
computed without gradient from the first flow draw $(\mathbf{z}_t,t)$ with scale $w=4$. Since the target velocity is constant along the path, $\hat{\mathbf{z}}$ is the current estimate of the clean code from any intermediate state, and mirrors the thoughts produced at inference. Note that it costs two extra forward passes per row, one conditioned and one with the null question.

\textbf{Question masking.}
The answer pass always sees the question, and the null-token replacement for classifier-free guidance applies to the flow pass only.

\subsubsection{Noise Draws per Code}
\label{app:draws}

\begin{figure}[ht]
    \centering
    \includegraphics[width=0.75\linewidth]{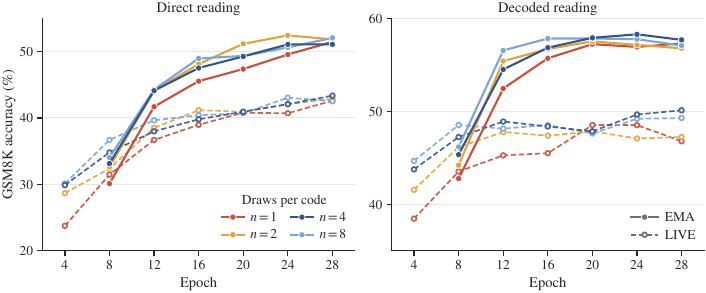}
    \caption{Noise draws per code. GSM8K accuracy (\%) over training for one, two, four and eight draws per code, under the direct and decoded readings, for the EMA weights (solid, filled markers) and the LIVE weights (dashed, open markers).}
    \label{fig:draws-curve}
\end{figure}

Instead of a single noise draw per example, the flow loss of \eqref{eq:fm} is computed over $n$ independent draws of $(t,\boldsymbol{\epsilon})$ for the same latent code and averaged as in MAR~\citep{li2024mar}. \figref{fig:draws-curve} tracks the accuracy of the EMA and LIVE weights over training for $n$ from 1 to 8.
With EMA weights, the final accuracies remain within about one point across all numbers of draws on both readings. Multiple draws mainly improve convergence speed. With a single draw, the EMA weights trail the other three by 2 to 4 points on the direct reading from epoch 8 to 20 and on the decoded reading at epoch 12, catching up by epochs 28 and 20, respectively, while two, four and eight draws stay within about two points of each other throughout. The LIVE weights show the same gap earlier, with a single draw 5 to 6 points behind on the direct reading and 3 to 6 on the decoded one at epoch 4, closing by epoch 20, and they end 7 to 10 points below the EMA weights. However, note that each draw costs one more flow pass per row, so that eight draws cost about 1.6 times the runtime of four, and we keep $n=4$.

\subsection{Flow Model, Stage 2}
\label{app:stage2}

Stage 2 is the self-training loop of \secref{subsec:flow-training} run in blocks of $B=16$ updates. At the start of each block, the EMA model proposes $K$ thoughts for each of the $N$ new questions, and the frozen decoder converts each thought into a symbolic CoT. CoTs whose final answer matches the reference are retained and re-encoded by the VAE. The next $B$ updates use these codes as flow targets and backpropagate an answer loss through a fresh rollout. The block is then replaced by the next one and the loop cycles over the question pool. We describe each step in the order of the loop next.

\textbf{Question pool.}
The pool holds 100K questions of OMI-2~\citep{toshniwal2025openmathinstruct2} with their reference answers and no CoT, disjoint from $\mathcal{D}_{\mathrm{flow}}$ and from the GSM8K test set. A seeded permutation of the pool is split across the 16 GPUs, and each block takes the next $N=1{,}024$ questions, 64 per GPU. The permutation wraps around, so a question left unsolved is proposed again in the next pass.

\textbf{Proposals.}
The proposer is the EMA of the trained weights (decay $0.999$) updated after every optimizer step. For each question, it draws $K=8$ Gaussian samples $\boldsymbol{\eta}_k\sim\mathcal{N}(\mathbf{0},\mathbf{I})$ and centers and rescales them into the starts
\[
\boldsymbol{\epsilon}_k=\sqrt{\frac{K}{K-1}}\left(\boldsymbol{\eta}_k-\frac{1}{K}\sum_{j=1}^{K}\boldsymbol{\eta}_j\right),
\]
so that each start is still a standard Gaussian while the $K$ starts sum to zero. Each start is then integrated as at inference, with $S=20$ Euler steps on the uniform grid $t_s=s/S$ and the guided velocity of scale $w=4$,
\begin{equation}
\begin{aligned}
\tilde{\mathbf{v}}_\theta(\mathbf{z},t,q)&=\mathbf{v}_\theta(\mathbf{z},t,\emptyset)+w\,\big[\mathbf{v}_\theta(\mathbf{z},t,q)-\mathbf{v}_\theta(\mathbf{z},t,\emptyset)\big],\\
\mathbf{z}^{(s+1)}&=\mathbf{z}^{(s)}+(t_{s+1}-t_s)\,\tilde{\mathbf{v}}_\theta\big(\mathbf{z}^{(s)},t_s,q\big),
\end{aligned}
\label{eq:guided-euler}
\end{equation}
from $\mathbf{z}^{(0)}=\boldsymbol{\epsilon}_k$ at unit noise scale, to the thought $\hat{\mathbf{z}}_k=\mathbf{z}^{(S)}$, and we write this map as $G_\theta^{S}(q,\boldsymbol{\epsilon})=\hat{\mathbf{z}}$. Proposals use it with the EMA weights and without gradient, and the answer loss below uses it with the LIVE weights and with gradients.

\textbf{Why centered starts.}
Independent draws share a random common offset which pushes all $K$ thoughts of a question in the same direction. Centering removes this offset, and the rescaling keeps each start an exact sample of the Gaussian prior so a single proposal is as accurate as before. In a matched pilot on 10,000 training questions with $K=16$, centered and independent starts give the same share of correct candidates (93.1\%), but centered starts raise the median number of distinct decoded CoTs per question from one to two and find a new verified CoT for 5.3\% more questions.

\textbf{Verification and targets.}
Each thought is un-standardized and decoded greedily by the frozen decoder into a symbolic CoT, without the question and with at most 160 new tokens. The CoT is accepted when its final answer parses and matches the reference under a strict grader. Verification thus only compares the final answer of the decoded CoT with the reference answer, rather than the answer of the direct reading, and it does not certify the intermediate steps.
The accepted CoTs of a question are deduplicated on their step text and all distinct ones are kept, while unsolved questions are dropped. Each kept CoT $r$ is re-encoded by the frozen encoder into its posterior mean and standardized with the unchanged stage 1 statistics, $\bar{\mathbf{z}}=(\boldsymbol{\mu}_\phi(r)-\mathbf{m})/\mathbf{s}$. These codes are then used as flow targets and form the training block $\mathcal{B}=\{(q,\bar{\mathbf{z}},a)\}$. At the start of stage 2 with the EMA stage 1 parent, 65 to 66\% of the questions of a block are solved with 2.5 to 2.7 distinct CoTs each.

\textbf{Batches.}
Each of the next $B=16$ updates takes 4 rows per GPU from $\mathcal{B}$. Each GPU goes through its accepted questions in a shuffled order, reshuffling when they are exhausted. Each update also adds one replay row per GPU drawn uniformly with replacement from $\mathcal{D}_{\mathrm{flow}}$ and encoded on the fly. After $B$ updates, the block is replaced even if some of its CoTs were not used.

\textbf{Losses.}
The flow loss of \eqref{eq:fm} is computed separately on the online rows of $\mathcal{B}$ and on the replay rows. The two are then mixed as
\begin{equation}
\mathcal{L}_{\mathrm{FM}}^{\mathrm{S2}}=(1-\alpha)\,\mathcal{L}_{\mathrm{FM}}^{\mathcal{B}}+\alpha\,\mathcal{L}_{\mathrm{FM}}^{\mathcal{D}_{\mathrm{flow}}},
\qquad \alpha=0.25 .
\label{eq:stage2-fm}
\end{equation}
The answer loss reads a thought generated by the LIVE weights from fresh noise over $S_{\mathrm{train}}=10$ guided Euler steps of \eqref{eq:guided-euler}, with gradients through every step and both guidance branches,
\begin{equation}
\mathcal{L}_{\mathrm{roll}}
=\mathbb{E}_{(q,a)\sim\mathcal{B},\,\boldsymbol{\epsilon}\sim\mathcal{N}(\mathbf{0},\mathbf{I})}
\Big[-\log p_\theta\big(a\mid q,G_\theta^{S_{\mathrm{train}}}(q,\boldsymbol{\epsilon})\big)\Big],
\label{eq:stage2-rollout}
\end{equation}
where the thought sits at the slot positions at $t=1$ with the question, as in stage 1, and $a$ is the reference answer. This rollout starts from new independent noise and not from the proposal starts. The loss is averaged over the answer tokens and is not applied to replay rows.
The objective is
\begin{equation}
\mathcal{L}_{\mathrm{Stage\,2}}=\lambda_{\mathrm{FM}}\,\mathcal{L}_{\mathrm{FM}}^{\mathrm{S2}}+\mathcal{L}_{\mathrm{roll}},
\qquad \lambda_{\mathrm{FM}}=5,
\label{eq:stage2}
\end{equation}
in which each term has one role: the online flow loss reproduces the verified thoughts, the replay term keeps the flow close to the stage 1 data, and the answer loss is the only term backpropagated through the full rollout.

\textbf{Optimization and evaluation.}
Stage 2 starts from the EMA weights of the stage 1 parent with a fresh AdamW optimizer ($\epsilon=10^{-6}$, other settings in \tabref{tab:defaults}) and runs for $1{,}600$ updates with seed 42. This corresponds to 100 blocks and 100K question visits, approximately one pass over the pool, though not over every retained CoT. Unlike stage 1, the final stage 2 model is the LIVE model, not the EMA. On 16 H100 GPUs, a block takes 130 to 160\,s to propose and verify and an update 4 to 5\,s, about 6 hours per run.

\textbf{Ablations.}
The ablations of \tabref{tab:stage2} keep everything else fixed and change one part of the loop: the targets, the flow loss or the answer loss, as follows.
\begin{itemize}[leftmargin=*,itemsep=1pt,topsep=2pt,parsep=0pt]
\item \emph{No verification} drops only the answer check: parseable CoTs are still deduplicated, and the answer loss still uses the reference answer rather than the possibly wrong decoded one.
\item \emph{One trace} keeps a single accepted CoT per question instead of all distinct ones.
\item \emph{Raw latents} keeps verification and deduplication but trains on the first accepted generated endpoint of each distinct CoT instead of its re-encoded code.
\item \emph{No online flow loss} removes only $(1-\alpha)\mathcal{L}_{\mathrm{FM}}^{\mathcal{B}}$ and keeps the answer loss and the replay term $\lambda_{\mathrm{FM}}\alpha\mathcal{L}_{\mathrm{FM}}^{\mathcal{D}_{\mathrm{flow}}}$ unchanged to isolate the contribution of the verified flow targets.
\item \emph{No stage 1 replay} sets $\alpha=0$ so the online flow loss takes the full weight.
\item \emph{No answer loss} removes $\mathcal{L}_{\mathrm{roll}}$ and keeps both the online flow loss and the replay.
\item \emph{Detached} keeps the answer loss but stops its gradient at the generated thought to isolate the effect of backpropagating the gradients through the full flow rollout.
\item \emph{1-step rollout} and \emph{5-step rollout} generate the thought of $\mathcal{L}_{\mathrm{roll}}$ with $S_{\mathrm{train}}=1$ or 5 guided steps instead of 10.
\end{itemize}

\textbf{Results.}
Stage 2 improves the direct reading by 3.8 points and the decoded reading by 1.3 over stage 1 (\tabref{tab:stage2}), and these gains come from the answer loss through the full rollout. Without the answer loss, the model stays at the stage 1 level. Detaching the rollout endpoint loses 2.3 points on the direct reading and 2.6 on the decoded one, and a one-step rollout loses 2.8 and 1.9. Verification is also needed, as training on unverified thoughts loses 2.0 and 1.9 points. Raw generated latent targets, one verified trace per question, no online flow loss, no stage 1 replay and a five-step rollout all stay within one point of stage 2 on both readings.

\subsection{Variable-Length Latent Thoughts}
\label{app:varslots}

A fixed width asks one code to hold both a one-step and a long CoT, while explicit CoT spends more tokens on harder questions. Our recipe supports a latent thought of variable width with almost no change, and unlike LaDiR~\citep{kang2025ladir}, which varies the length by generating one block after another, the thought stays a single block of slots refined in parallel and only gets wider or narrower. This keeps the method as efficient as with a fixed width since the number of sequential passes stays $2S$ whatever the width, whereas block-wise generation repeats the whole denoising for every block. The VAE registers the largest number of slot tokens and uses the first $M$ of them, the flow model generates $M$ slots with the loss of \eqref{eq:fm} averaged over the active slots, and $M$ can be chosen at inference.

\textbf{Two designs.}
The VAE has 16 slot tokens and encodes each CoT with 8, 12 or 16 of them, chosen by the length of the CoT: 8 up to 16 encoder tokens, 12 up to 24, and 16 beyond. In a quarter of the cases one of the two other widths is used instead, so that every width is trained on CoTs of every length, and the flow model is trained on the same mix of widths. The two designs differ in how the codes of different widths relate. In the \emph{nested} design, each slot token attends only to the slots before it, so the first 8 slots of a 16-slot code are exactly the 8-slot code, and one encoder pass at 16 slots gives all three widths, as in Matryoshka representations~\citep{rippel2014nested,kusupati2022matryoshka} and nested tokenizers~\citep{bachmann2025flextok}. In the \emph{elastic} design, the slot tokens attend to each other in both directions, so each width has its own code, computed by a separate encoder pass per width, as in elastic tokenizers~\citep{yan2024elastictok}.

\textbf{Outcome on GSM8K-Aug.}
The nested design trains and generates at every width as intended, but on GSM8K-Aug it brings no gain since 8 slots already suffice. The symbolic CoTs are short, with a median of 21 encoder tokens and 90\% below 35 tokens. Eight slots already reconstruct 98.7\% of the test CoTs exactly, and increasing the number of slots to 16 adds only 0.4 points. The extra slots therefore carry little, a weak residual of the first eight at about half their scale. As a result, the best variable-width model only matches the fixed width of 8 (60.1 decoded for both), and choosing the width by the length of the CoT does not beat it. While variable widths do not help on this dataset, we expect them to on data whose CoTs vary more in length, since the recipe is flexible and applies to variable widths with almost no change. In that case, the width becomes a thinking budget chosen at inference~\citep{muennighoff2025s1,aggarwal2025l1}.

\subsection{Baseline Numbers}
\label{app:baselines}

For each baseline entry in \tabref{tab:main}, we use the result from the original paper when it reports the relevant backbone and test set. Otherwise, we use our own run or evaluate released weights, leaving the entry empty when neither is available. \tabref{tab:baseline-sources} lists the sources. We do not copy rows from a single reproduction, since reproductions of one baseline differ by up to nine points, \eg{}, explicit CoT on Qwen2.5-0.5B-Instruct reaches 50.6 in \citet{kuzina2026kava} and 59.7 in our run.

\begin{table}[ht]
\caption{Source of each baseline number of \tabref{tab:main}. P: printed by the paper that introduced the method, CODI~\citep{shen2025codi}, PCCoT~\citep{wu2025pccot} or KaVa~\citep{kuzina2026kava}. T: trained by us, with LaDiR reimplemented as in Appendix~\ref{app:ladir-baseline}. R: released weights evaluated by us, the public explicit CoT checkpoint \texttt{yingfanbot/gsm-cot-llama3b} (a), and those of \citet{dilgren2026interpretable} (b) and \citet{wu2025pccot} (c).}
\label{tab:baseline-sources}
\vspace{4pt}
\centering
\footnotesize\setlength{\tabcolsep}{5pt}
\begin{tabular}{@{}llll@{}}
\toprule
Method & Qwen2.5-0.5B-Instruct & Llama-3.2-1B-Instruct & Llama-3.2-3B-Instruct \\
\midrule
Explicit CoT & T & T & R (a) \\
No-CoT & T & T & T \\
iCoT & T & T & T \\
Coconut & T & R (b) & T \\
CODI & T & P & T \\
PCCoT & T & P on GSM8K, R (c) on the OOD sets & T \\
KaVa & P, no MultiArith & P, no MultiArith & P, no MultiArith \\
LaDiR & T & T & T \\
\bottomrule
\end{tabular}
\end{table}

\begin{table}[ht]
\caption{Recipes of our baseline runs on GSM8K-Aug. Learning rates are for Qwen2.5-0.5B / Llama-3.2-1B / Llama-3.2-3B.}
\label{tab:baseline-recipes}
\vspace{4pt}
\centering
\footnotesize\setlength{\tabcolsep}{3pt}
\begin{tabular}{@{}llll>{\raggedright\arraybackslash}p{6.3cm}@{}}
\toprule
Method & Tuning & Learning rate & Epochs & Other settings \\
\midrule
Explicit CoT & LoRA, $r=128$, $\alpha=32$ & $8\cdot10^{-4}$ / $8\cdot10^{-4}$ / -- & 10 & batch 128, cosine, 3\% warmup, wd 0.1 \\
No-CoT & LoRA, $r=128$, $\alpha=32$ & $2\cdot10^{-4}$ / $8\cdot10^{-4}$ / $2\cdot10^{-4}$ & 10 & as explicit CoT, answer only \\
iCoT & full, single precision & $10^{-5}$ & 20 & batch 32, 8 CoT tokens removed per epoch (Qwen: 11) \\
Coconut & full, bf16 & $5\cdot10^{-5}$ & 3 + 6 + 1 & $c=1$, a CoT stage, stages 1 to 6 and a fully latent stage, batch 128 \\
CODI & LoRA, $r=128$, $\alpha=32$ & $8\cdot10^{-4}$ / -- / $3\cdot10^{-4}$ & 10 / -- / 8 & 6 latents, distillation weight 20, batch 128 \\
PCCoT & LoRA, $r=128$, $\alpha=32$ & $5\cdot10^{-4}$ / -- / $2\cdot10^{-4}$ & 10 & 24 latents, $T=3$ iterations, batch 128 \\
\bottomrule
\end{tabular}
\end{table}

\textbf{Our runs.}
We train each method with its official code patched only for Qwen2 support (PCCoT and Coconut) and for multi-GPU training and greedy decoding (iCoT), with the recipes of \tabref{tab:baseline-recipes}. Where the original paper has no recipe for a backbone, we follow the closest reproduction: \citet{kuzina2026kava} for the LoRA baselines and PCCoT, \citet{wei2025simcot} for CODI at 3B, and the 1B recipe of \citet{dilgren2026interpretable} for Coconut.

\textbf{Evaluation.}
We evaluate both our runs and released weights with the same protocol as our model (\tabref{tab:defaults}), using greedy decoding on the full test sets and matching the last number in each output to the reference answer. This protocol agrees with that of the original papers: on the released 1B checkpoints, we obtain 55.6 on GSM8K for CODI against the printed 55.6, and 54.1 for PCCoT against 53.35.

\subsection{LaDiR Baseline}
\label{app:ladir-baseline}

We reimplement LaDiR~\citep{kang2025ladir} as a variable-length, block-autoregressive latent baseline, trained on the same GSM8K-Aug and Diverse CoT rows as our model.

\textbf{VAE.}
Each symbolic step takes the place of a sentence and is encoded into its own block, a code of three slots of dimension 512 (so that, averaged over the dataset, a CoT takes about as many slots as our fixed 8). In the last block we include an answer-only suffix so that a stopping head can learn when the CoT is complete. The encoder and decoder (Llama-3.2-1B and Llama-3.2-3B) start from the OMI-2 language VAE that also initializes ours, and the encoder is fine-tuned, on single steps, with the objective of \eqref{eq:vae}, with token substitution and additive Gaussian latent noise.

\textbf{Flow model.}
The flow model uses the same backbones as ours, Llama-3.2-1B-Instruct, Llama-3.2-3B-Instruct or Qwen2.5-0.5B-Instruct. In stage 1 it predicts the velocity of each noised block given the clean preceding blocks, with uniform time and bidirectional attention within each block, and the answer and stopping losses read the clean codes. Stage 2 keeps the same supervised rows but replaces the preceding blocks with blocks generated by 10-step guided rollouts, with gradients through the rollout. The flow, answer and stopping losses are weighted 5, 1 and 2, and the optimizer is that of our recipe (\tabref{tab:defaults}).

\textbf{Inference.}
Blocks are generated sequentially until the stopping head fires or 16 blocks have been generated, using guidance $w=4$ and noise scale 2. The answer is then sampled at temperature 0.7. As for our model, each block takes 20 Euler steps, and stage 1 is read with the EMA weights and stage 2 with the LIVE weights.

\textbf{Efficiency comparison.}
LaDiR generates one block per reasoning step, and each block needs a full denoising of $S$ guided steps, as much as our whole thought. Its cost therefore grows with the length of the reasoning. GSM8K test CoTs have 3.3 steps on average and 5 at the 90th percentile, so at 20 steps LaDiR runs about 130 sequential forward passes per question, against 40 for our model. At our measured 22\,ms per guided Euler step, this is about 1.5\,s per question on average and 2.2\,s at the 90th percentile, against 0.46\,s for our direct reading, or 3.2 to 4.8 times slower than our model and 5.5 times slower than explicit CoT. The gap widens with fewer steps. LaDiR needs 10 to 20 Euler steps per block, since its answer is not trained on the model's own one-step endpoints and each block builds on the previously generated ones, whose errors accumulate. Our answer pass is trained to read such endpoints in stage 1, and our direct reading reaches 57.4 after only two steps, in 68\,ms, 11 to 21 times faster than LaDiR at 10 to 20 steps per block.

\section{Algorithms}
\label{app:algorithms}

Algorithms~\ref{alg:stage1}, \ref{alg:stage2} and~\ref{alg:inference} outline stage 1, stage 2 and inference. The VAE, trained beforehand with \eqref{eq:vae-dual}, stays frozen, $\tilde{\mathbf{v}}_\theta(\mathbf{z},t,q)=\mathbf{v}_\theta(\mathbf{z},t,\emptyset)+w\,[\mathbf{v}_\theta(\mathbf{z},t,q)-\mathbf{v}_\theta(\mathbf{z},t,\emptyset)]$ is the guided velocity with scale $w$, and $\operatorname{sg}[\cdot]$ stops the gradient.

\begin{algorithm}[H]
\caption{Stage 1 training of the flow model}
\label{alg:stage1}
\small
\begin{algorithmic}[1]
\Require data $\mathcal{D}=\{(q,r_{\mathrm{sym}},a)\}$, frozen VAE encoder $\phi$ with statistics $(\mathbf{m},\mathbf{s})$, flow model $\theta$
\Repeat
  \State sample $(q,r_{\mathrm{sym}},a)\sim\mathcal{D}$
  \State $\bar{\mathbf{z}}\gets(\boldsymbol{\mu}_\phi(r_{\mathrm{sym}})-\mathbf{m})/\mathbf{s}$ \Comment{encode the CoT}
  \State $\boldsymbol{\epsilon}\sim\mathcal{N}(\mathbf{0},\mathbf{I})$, \ $t\sim\text{logit-normal}(-2,0.8^2)$ \Comment{shifted toward noise}
  \State $\mathbf{z}_t\gets(1-t)\,\boldsymbol{\epsilon}+t\,\bar{\mathbf{z}}$ \Comment{point on the straight path from noise to code}
  \State $q'\gets\emptyset$ with probability 0.1, else $q$ \Comment{null question for CFG}
  \State $\mathcal{L}_{\mathrm{FM}}\gets\big\|\mathbf{v}_\theta(\mathbf{z}_t,t,q')-(\bar{\mathbf{z}}-\boldsymbol{\epsilon})\big\|_2^2$ \Comment{flow pass}
  \If{with probability 0.5} \Comment{input of the answer pass, 50/50}
    \State $\mathbf{z}_{\mathrm{ans}}\gets\operatorname{sg}\big[\mathbf{z}_t+(1-t)\,\tilde{\mathbf{v}}_\theta(\mathbf{z}_t,t,q)\big]$ \Comment{detached one-step endpoint}
  \Else
    \State $\mathbf{z}_{\mathrm{ans}}\gets(1-\rho)\,\bar{\mathbf{z}}+\rho\,\boldsymbol{\epsilon}$, \ $\rho\sim\mathcal{U}[0,0.3]$ \Comment{noised code}
  \EndIf
  \State $\mathcal{L}_{\mathrm{CE}}\gets-\log p_\theta(a\mid q,\mathbf{z}_{\mathrm{ans}})$ \Comment{answer pass}
  \State gradient step on $\lambda_{\mathrm{FM}}\mathcal{L}_{\mathrm{FM}}+\mathcal{L}_{\mathrm{CE}}$, then update the EMA $\bar\theta$
\Until{the end of training}
\Ensure EMA weights $\bar\theta$
\end{algorithmic}
\end{algorithm}

\begin{algorithm}[H]
\caption{Stage 2 training of the flow model}
\label{alg:stage2}
\small
\begin{algorithmic}[1]
\Require stage 1 EMA weights $\theta$ and data $\mathcal{D}$, frozen VAE $(\phi,\psi)$ with $(\mathbf{m},\mathbf{s})$, question pool $\mathcal{Q}=\{(q,a)\}$
\State $\bar\theta\gets\theta$ \Comment{EMA proposer}
\For{each block of $N$ new questions from $\mathcal{Q}$}
  \For{each $(q,a)$ in the block, $K$ times} \Comment{propose and verify, without gradient}
    \State $\hat{\mathbf{z}}\gets\Call{Think}{\bar\theta,q,S}$ \Comment{Algorithm~\ref{alg:inference}}
    \State decode $\hat r$ from $p_\psi(\cdot\mid\mathbf{m}+\mathbf{s}\odot\hat{\mathbf{z}})$ \Comment{symbolic route}
    \If{the final answer of $\hat r$ is $a$} \Comment{verification}
      \State add $\big(q,\,(\boldsymbol{\mu}_\phi(\hat r)-\mathbf{m})/\mathbf{s},\,a\big)$ to the targets $\mathcal{B}$ of the block \Comment{re-encode as a flow target}
    \EndIf
  \EndFor
  \For{$B$ updates} \Comment{train on the block}
    \State sample $(q,\bar{\mathbf{z}},a)\sim\mathcal{B}$ and a replay row from $\mathcal{D}$
    \State $\mathcal{L}_{\mathrm{FM}}\gets(1-\alpha)\,\mathcal{L}_{\mathrm{FM}}^{\mathcal{B}}+\alpha\,\mathcal{L}_{\mathrm{FM}}^{\mathcal{D}}$ \Comment{flow loss of Algorithm~\ref{alg:stage1} on each row}
    \State $\hat{\mathbf{z}}\gets\Call{Think}{\theta,q,S_{\mathrm{train}}}$ \Comment{rollout from fresh noise, with gradients}
    \State $\mathcal{L}_{\mathrm{roll}}\gets-\log p_\theta(a\mid q,\hat{\mathbf{z}})$ \Comment{answer loss through the full rollout}
    \State gradient step on $\lambda_{\mathrm{FM}}\mathcal{L}_{\mathrm{FM}}+\mathcal{L}_{\mathrm{roll}}$, then update the EMA $\bar\theta$
  \EndFor
\EndFor
\Ensure LIVE weights $\theta$
\end{algorithmic}
\end{algorithm}

\begin{algorithm}[H]
\caption{Inference, with the decoded CoTs as an optional probe}
\label{alg:inference}
\small
\begin{algorithmic}[1]
\Require question $q$, flow model $\theta$, Euler steps $S$, guidance scale $w$, VAE decoder $\psi$ with $(\mathbf{m},\mathbf{s})$
\Function{Think}{$\theta,q,S$}
  \State $\mathbf{z}\sim\mathcal{N}(\mathbf{0},\mathbf{I})$ \Comment{start from noise at $t=0$}
  \For{$i=0,\dots,S-1$}
    \State $\mathbf{z}\gets\mathbf{z}+\tilde{\mathbf{v}}_\theta(\mathbf{z},i/S,q)/S$ \Comment{guided Euler step}
  \EndFor
  \State \Return $\mathbf{z}$ \Comment{latent thought at $t=1$}
\EndFunction
\State $\hat{\mathbf{z}}\gets\Call{Think}{\theta,q,S}$
\State generate $a$ from $p_\theta(\cdot\mid q,\hat{\mathbf{z}})$ \Comment{answer, the direct reading}
\Statex \textit{Optional probe: read the thought back as an explicit CoT}
\State $\mathbf{z}\gets\mathbf{m}+\mathbf{s}\odot\hat{\mathbf{z}}$ \Comment{undo the standardization}
\State decode $\hat r_{\mathrm{sym}}$ from $p_\psi(\cdot\mid\mathbf{z})$ \Comment{symbolic route, the decoded reading}
\State decode $\hat r_{\mathrm{lang}}$ from $p_\psi(\cdot\mid\mathbf{z},q)$ \Comment{language route, given the question}
\Ensure answer $a$, and optionally the CoTs $\hat r_{\mathrm{sym}}$ and $\hat r_{\mathrm{lang}}$
\end{algorithmic}
\end{algorithm}

\section{Datasets}
\label{app:data}

\tabref{tab:app-data} lists the corpora used in the paper.

\begin{table}[ht]
\caption{Corpora used in the paper. Rows are (question, CoT, answer) triples and questions are distinct normalized question texts.}
\label{tab:app-data}
\begin{center}
\small
\begin{tabular}{@{}llrr@{}}
\toprule
\textbf{Corpus} & \textbf{Origin} & \textbf{Rows} & \textbf{Questions} \\
\midrule
GSM8K-Aug, flow & \texttt{whynlp/gsm8k-aug} train & 385K & 385K \\
GSM8K-Aug, paired & aligned with \texttt{whynlp/gsm8k-aug-nl} & 367K & 367K \\
Diverse CoT & Qwen2.5-32B-Instruct on GSM8K-Aug questions & 216K & 148K \\
Diverse CoT, paired & same, with a verified natural language twin & 205K & 142K \\
IID data & OMI-2 \texttt{augmented\_gsm8k}, converted & 166K & 39K \\
OOD data & OMI-2 \texttt{augmented\_math}, converted & 463K & 80K \\
Stage 2 pool & OMI-2 questions and answers, no CoT & 100K & 100K \\
\bottomrule
\end{tabular}
\end{center}
\end{table}

\textbf{GSM8K-Aug and GSM8K-Aug-NL.}
GSM8K-Aug~\citep{deng2023implicit} has 385K training rows, nearly one per distinct question, 500 validation rows and the 1,319 GSM8K test questions~\citep{cobbe2021gsm8k}.
Its CoT is symbolic, one marker \texttt{<<expression=value>>} per step, such as \texttt{<<600*30/100=180>>}, and the last value is the answer.
GSM8K-Aug-NL writes the same steps of the same questions as sentences.
We pair the two releases row by row into $\mathcal{D}_{\mathrm{vae}}$ and keep for the VAE the 367K training pairs whose symbolic chain ends at the answer, whose last sentence contains it, and whose intermediate values all appear in the sentences.

\textbf{Test sets.}
We evaluate on the 1,319 GSM8K test questions and, zero-shot, on GSM8K-Hard~\citep{gao2023pal}, the same 1,319 questions with larger numbers, on the 1,000 questions of SVAMP~\citep{patel2021svamp}, and on the 180 test questions of MultiArith~\citep{roy2015multiarith}.

\textbf{Diverse CoT.}
Qwen2.5-32B-Instruct~\citep{qwen2025qwen25} wrote 16 solutions per GSM8K-Aug question directly in the GSM8K-Aug format. A solution is kept when it parses, computes correctly, ends at the reference answer, differs from the stored solution and passes two semantic judges, which leaves 216K CoTs over 148K questions.

\textbf{IID and OOD data.}
Both come from OpenMathInstruct-2 (OMI-2)~\citep{toshniwal2025openmathinstruct2}, whose CoTs are in natural language and are converted to symbolic CoTs by the pipeline of Appendix~\ref{app:converter}. IID data is its \texttt{augmented\_gsm8k} share and OOD data its \texttt{augmented\_math} share, with no question overlapping any split of GSM8K-Aug.
The VAE data of \tabref{tab:vae-data} adds the paired rows of one source to $\mathcal{D}_{\mathrm{vae}}$ (572K, 533K, 830K and 996K rows), and the flow data of \tabref{tab:flow-data} adds its symbolic rows to the 385K GSM8K-Aug rows (602K, 551K, 849K and 1,015K rows).

\textbf{Training data of the VAE.}
\label{app:vae-data}
The default VAE encodes the symbolic CoTs of GSM8K-Aug, which are of limited diversity with mostly one CoT per question. We thus ask whether a more diverse corpus induces a better latent space, with the flow data held fixed, by adding the Diverse CoT, the IID data or the OOD data above. As shown in \tabref{tab:vae-data}, more data can help but only to a very limited degree, since reconstruction is already saturated on the default corpus.

\begin{table}[ht]
\caption{Training data of the VAE, with the flow data fixed. Reconstruction (\%) and GSM8K accuracy (\%) under the direct and decoded readings.}
\label{tab:vae-data}
\vspace{4pt}
\centering
\small\setlength{\tabcolsep}{3pt}
\begin{tabular}{@{}lccc@{}}
\toprule
\textbf{VAE training data} & \textbf{Recon.} & \textbf{Direct} & \textbf{Decoded} \\
\midrule
GSM8K-Aug CoT & 99.1 & 50.7 & 56.9 \\
+ Diverse CoT & 99.0 & 48.8 & 54.8 \\
+ IID data & 99.0 & \textbf{51.6} & 56.9 \\
+ OOD data & 99.3 & 50.9 & 56.0 \\
+ IID and OOD data ($5\times$ budget) & \textbf{99.6} & 51.1 & \textbf{58.3} \\
\bottomrule
\end{tabular}
\end{table}

\textbf{Stage 2 pool.}
The 100K OMI-2 questions of Appendix~\ref{app:stage2} are 73K \texttt{augmented\_gsm8k} questions and 26K \texttt{augmented\_math} questions with their reference answers only and no CoT.

\section{Symbolic Conversion Pipeline}
\label{app:converter}

This pipeline turns a dataset with natural language CoTs into the paired format of $\mathcal{D}_{\mathrm{vae}}$.

\textbf{The symbolic language.}
The converter targets a small-step language that includes GSM8K-Aug arithmetic as a subset. Each CoT consists of \texttt{<<BODY>>} markers in ASCII mathematics, with one move per marker: an equation, an inequality, a declaration \texttt{<<x in R>>}, a definition \texttt{<<u\_1:=3*4>>}, or a guarded case. Every name must come from the question or an earlier declaration, and the final marker contains only the answer.

\textbf{Converter.}
Qwen3.8-27B~\citep{qwen2026qwen38}, sampled at temperature 1 with its highest reasoning effort and a fixed seed per row, is given the language and its rules, the question, the CoT split into numbered sentences, and the reference answer. It returns groups of elementary symbolic steps, each naming the sentences it compiles, or rejects the CoT, or flags the question as ill-posed, and the natural language CoT is never rewritten.

\textbf{Run on OMI-2.}
Before conversion, a filter keeps 2M of the 13M rows of OMI-2. It rejects rows with a false equation or a conclusion that contradicts the label, keeps CoTs under 256 tokens with explicit reasoning and no backtracking, removes the questions of all GSM8K-Aug sets, and keeps at most eight distinct solutions per question. A blind audit, in which the converter answers each question alone, then keeps the 57.5\% of the questions whose answer matches the label. The converter succeeds on 96.3\% of the remaining rows, and a review in a fresh context, which sees the question, the original CoT and the symbolic one together, accepts 90.5\% of the 700,131 distinct traces. After deduplication and a final leakage check, 629,824 pairs over 119,835 questions remain.

\textbf{Examples.}
\tabref{tab:converter-examples} shows two published pairs, one IID and one OOD.
\begin{table}[ht]
\caption{Two published pairs, one IID and one OOD, with the natural language CoT verbatim from OMI-2 and the symbolic CoT written by the converter.}
\label{tab:converter-examples}
\vspace{2pt}
\centering\footnotesize
\newcommand{\symstep}[1]{\mbox{\fontfamily{lmtt}\selectfont\textcolor{black!40}{<<}#1\textcolor{black!40}{>>}}}
\begin{tabular}{@{}p{0.11\linewidth}p{0.855\linewidth}@{}}
\toprule
\multicolumn{2}{@{}l}{\emph{IID pair, from the {\fontfamily{lmtt}\selectfont augmented\_gsm8k} share}} \\
\arrayrulecolor{black!20}\midrule
Question & Lucy has orange trees that produce 4 oranges each. If an orange can be sold for \$2 and Lucy wants to earn \$80, how many orange trees does she need to harvest? \\
\midrule
$r_{\mathrm{lang}}$ & Each orange tree produces 4 oranges. Each orange can be sold for \$2. So the amount of money Lucy earns from each tree is 4 * \$2 = \$8. If Lucy wants to earn \$80, then she needs to harvest \$80 / \$8 = 10 trees. Thus, Lucy needs to harvest 10 orange trees. \\
\midrule
$r_{\mathrm{sym}}$ & \raggedright \symstep{4*2=8} \symstep{80/8=10} \symstep{10} \tabularnewline
\midrule
$a$ & 10 \\
\arrayrulecolor{black}\midrule
\multicolumn{2}{@{}l}{\emph{OOD pair, from the {\fontfamily{lmtt}\selectfont augmented\_math} share}} \\
\arrayrulecolor{black!20}\midrule
Question & The greatest common divisor of $a$, $12$, and $21$ is $3$. What is the smallest possible value of $a$ that is greater than $3$? \\
\midrule
$r_{\mathrm{lang}}$ & The greatest common divisor (GCD) of $12$ and $21$ is $3$. Since $a$ shares the same GCD, $a$ must be a multiple of $3$. The smallest multiple of $3$ greater than $3$ is $6$. Therefore, the smallest possible value of $a$ that is greater than $3$ is $6$. \\
\midrule
$r_{\mathrm{sym}}$ & \raggedright \symstep{a in Z} \symstep{gcd(12,21)=3 and gcd(a,12,21)=3=>rem(a,3)=0} \symstep{a>3 and rem(a,3)=0=>a>=6} \symstep{gcd(6,12,21)=3} \symstep{6} \tabularnewline
\midrule
$a$ & 6 \\
\arrayrulecolor{black}\bottomrule
\end{tabular}
\end{table}

\textbf{Future directions.}
While we provide this conversion as a simple offline LLM-based step, the converter could also be trained with execution feedback by writing symbolic CoTs as executable programs~\citep{gao2023pal,chen2023pot}, and the symbolic structure could even be induced directly in the latent space from task feedback alone~\citep{macfarlane2026nli}. We leave both directions for future work.

\section{The Requirements of Latent Reasoning: Details}
\label{app:properties}

All probes use Llama-3.2-1B-Instruct fine-tuned on GSM8K-Aug and, unless noted, the 1,319 GSM8K test questions. The baselines are the released checkpoints of CODI~\citep{shen2025codi} and PCCoT~\citep{wu2025pccot}, and those of Coconut, explicit CoT and No-CoT from \citet{dilgren2026interpretable}, whose reported accuracies we reproduce within half a point.
Our models run at 20 Euler steps and guidance 4, with one fixed noise draw per question so that every probe reads the same thought.

\subsection{Useful: The Twin Test}
\label{app:useful}

\begin{table}[h]
\centering\small
\caption{Twin test on 824 pairs. Share of pairs (\%) whose answer follows the injected thought (the twin's answer), keeps the original answer, or is neither. Clean follow: the share that follows, counted only on the pairs where the model answers both the question and its twin correctly without any injection.}
\label{tab:app-twin}
\setlength{\tabcolsep}{4pt}
\begin{tabular}{lrrrr}
\toprule
Model & Follows & Keeps original & Neither & Clean follow \\
\midrule
Explicit CoT (reference steps) & 97.2 & 0.2 & 2.5 & 99.9 \\
CODI & 31.8 & 19.8 & 48.4 & 68.4 \\
PCCoT & 27.7 & 20.8 & 51.6 & 62.3 \\
Coconut & 7.3 & 29.5 & 63.2 & 19.2 \\
\ours, stage 1 & \textbf{39.7} & \textbf{8.7} & 51.6 & \textbf{91.4} \\
\ours, stage 2 & 37.3 & 15.0 & 47.7 & 76.1 \\
\bottomrule
\end{tabular}
\end{table}

\textbf{Probe.}
The twin test asks whether the answer comes from the thought or from the question alone. Each test question gets a twin that changes one number, and we obtain the twin's CoT and answer by recomputing the reference CoT with the new number. For example, the twin of ``If Ann is 9 years old and her brother is twice her age, how old will her brother be in 3 years?'' makes Ann 8, so the CoT \texttt{<<9*2=18>> <<18+3=21>>} becomes \texttt{<<8*2=16>> <<16+3=19>>} and the answer changes from 21 to 19.
We first run the model on the twin and keep its thought, then run it on the original question with this thought in place of its own. The question now leads to 21 and the thought to 19: the answer follows the thought if it is 19, keeps the question if it is 21, and is neither otherwise. The injected thought is the twin's recomputed CoT for explicit CoT, and for ours the thought that the flow generates for the twin from the same noise draw.
We only change a number that appears once in the question and enters a single step, and keep a twin only if all step results remain whole numbers and its new number and answer do not appear in the original CoT, which gives 824 pairs.
A model can only follow a thought that leads to the twin's answer, so we also report the clean follow, the follow rate on the pairs where the model, without any injection, answers both the question and its twin correctly. There the injected thought is known to be right, and a model that uses its thought should almost always follow it.

\textbf{Results.}
Results are shown in \tabref{tab:app-twin}. Explicit CoT follows the injected steps on 97\% of the pairs, as a model that reasons through its steps should. CODI follows the injected thought on 32\%, PCCoT on 28\% and Coconut on 7\%, and even on the clean pairs CODI and PCCoT follow on about two thirds and Coconut on one fifth.
Our stage 1 and stage 2 models follow on 40\% and 37\% of all pairs, and on 91\% and 76\% of the clean pairs. When they do not follow, they rarely keep the original answer and mostly give neither, because the twin's thought is itself wrong: on 98\% and 96\% of the pairs with neither answer, our models also get the twin wrong without any injection.
Replacing the thought altogether shows the same dependence. The stage 1 model falls from 55.3 to 47.4 with the mean thought, to 40.6 with no thought and to 5.9 with the thought of another question. Stage 2 falls from 59.1 to 47.3, 39.3 and 21.6, relying a little more on the question after self-training.

\subsection{Diverse: Sampling by Noise}
\label{app:diverse}

\begin{figure}[ht]
    \centering
    \includegraphics[width=0.62\linewidth]{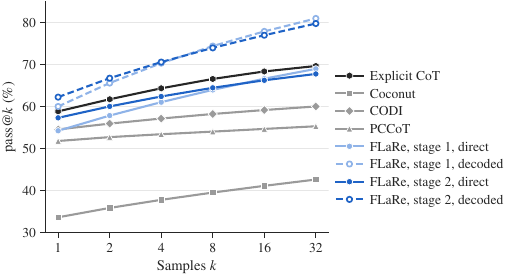}
    \caption{pass@$k$ on the GSM8K test set, estimated from 32 samples per question with the unbiased estimator of \citet{chen2021codex}. Our models generate flow samples from noise, the baselines add noise to their thought at the calibrated scale, and explicit CoT samples its tokens at temperature 0.7.}
    \label{fig:passk}
\end{figure}

\begin{table}[ht]
\centering\small
\caption{Sampling $K=16$ thoughts per question: accuracies (\%), distinct answers per question, and committed errors as a share of the greedy errors.}
\label{tab:app-sampling}
\setlength{\tabcolsep}{3.5pt}
\begin{tabular}{lrrrrrrr}
\toprule
Model & $\sigma$ & Greedy & pass@1 & pass@$K$ & Majority & Distinct ans. & Committed err. \\
\midrule
CODI & 0.5 & 55.6 & 54.6 & 59.1 & 54.9 & 1.46 & 51\% \\
PCCoT & 1.0 & 54.1 & 51.9 & 54.9 & 52.2 & 1.31 & 58\% \\
Coconut & 2.0 & 36.0 & 33.8 & 40.6 & 34.5 & 1.89 & 34\% \\
\ours, stage 1, decoded & -- & 61.2 & 60.1 & \textbf{77.8} & \textbf{64.4} & \textbf{3.41} & \textbf{6\%} \\
\ours, stage 1, direct & -- & 55.3 & -- & 67.1 & 55.6 & 2.68 & -- \\
\ours, stage 2, decoded & -- & \textbf{62.6} & \textbf{62.0} & 77.3 & \textbf{64.4} & 2.89 & 9\% \\
\ours, stage 2, direct & -- & 59.1 & -- & 66.2 & 58.2 & 1.93 & -- \\
\bottomrule
\end{tabular}
\end{table}

\textbf{Probe.}
We draw 16 samples per question. For ours, these use 16 noise draws of the flow, with the first serving as the greedy baseline. The baselines map a question to a single thought, so we add Gaussian noise to it, with a norm of $\sigma$ times that of the thought, at the largest $\sigma$ in $\{0.03, 0.1, 0.3, 0.5, 1, 2\}$ that keeps the single-sample accuracy within two points of greedy on the 500 validation questions.
A committed error is a greedy error that every sample repeats.

\textbf{Results.}
Results are shown in \tabref{tab:app-sampling} and \figref{fig:passk}. Noise changes the answer of the baselines on a fifth to a half of the questions, yet their pass@16 is at most 5 points above greedy, voting never beats greedy, and half of the errors of CODI and PCCoT are committed.
On the decoded reading, our stage 1 model reaches a pass@16 of 77.8 against a greedy 61.2, with 3.4 distinct answers per question, a 3.3-point gain from voting and only 6\% of greedy errors repeated by every sample, and stage 2 keeps most of this coverage.
The gap grows with more samples: from one to 32 samples, stage 2 gains 17.5 points, the baselines 3.5 to 9.0 and explicit CoT 10.8.

\subsection{Explainable: Reading the Thoughts}
\label{app:explainable}

\begin{table}[ht]
\begin{minipage}[b]{0.41\linewidth}
\caption{Values in order: questions (\%) whose reference intermediate results all appear in order, through the LM head for the baselines and in the decoded CoT for ours. Random: the same test with random values.}
\label{tab:app-explainable}
\vspace{4pt}
\centering\small\setlength{\tabcolsep}{3pt}
\begin{tabular}{@{}lrr@{}}
\toprule
Model & Values in order & Random \\
\midrule
CODI & 38.5 & 0.1 \\
Coconut & 19.1 & 0.2 \\
PCCoT & 7.8 & 0.2 \\
No-CoT, 6 pauses & 3.8 & 0.3 \\
\ours, stage 1 & 46.0 & 0.1 \\
\ours, stage 2 & \textbf{48.6} & 0.1 \\
\bottomrule
\end{tabular}
\end{minipage}\hfill
\begin{minipage}[b]{0.57\linewidth}
\caption{Faithfulness (\%) with the twin's thought injected on the pairs whose edit changes an intermediate value. Reads: the decoded thought shows the twin's values. Follows: the answer is the twin's. Reads and follows: both hold, so the answer matches what the thought shows.}
\label{tab:app-faithful}
\vspace{4pt}
\centering\small\setlength{\tabcolsep}{3pt}
\begin{tabular}{@{}lrrr@{}}
\toprule
& Reads & Follows & Reads and \\
Model & twin values & twin & follows \\
\midrule
CODI & 24.8 & 40.3 & 16.1 \\
PCCoT & 32.7 & 34.9 & 13.5 \\
Coconut & 44.7 & 8.9 & 6.4 \\
\ours, stage 1 & 81.0 & \textbf{44.1} & 40.5 \\
\ours, stage 2 & \textbf{81.5} & \textbf{44.1} & \textbf{41.8} \\
\bottomrule
\end{tabular}
\end{minipage}
\end{table}

\textbf{Probe.}
We decode each thought and check whether every intermediate result of the reference CoT appears in it in order, which we call values in order, with random values of the same length as the chance level. For ours, the frozen decoder turns the thought into a symbolic CoT, which must contain each result as an exact value, the final one included. The baselines have no decoder, so we read each thought through the model's own LM head and keep its top-1 token. A result counts as read when its first token matches one of these tokens, since Llama splits numbers above 999 into several tokens.
A No-CoT model read at six inserted pause positions gives the level reached without any thought.
For faithfulness, on the twin pairs whose edit changes an intermediate value, we read the injected thought in the same way, decoded for ours and through the LM head for the baselines, and check whether it shows the twin's values and whether the answer follows them.

\textbf{Results.}
Results are shown in \tabref{tab:app-explainable} and \tabref{tab:app-faithful}. The values are decoded in order for 46\% and 49\% of the questions with our stage 1 and stage 2 models, against 39\% for CODI, 19\% for Coconut and 8\% for PCCoT, although our match is stricter and counts a correct CoT that takes another route as a miss. All models stay above the No-CoT control (3.8\%) and far above chance (at most 0.3\%), so these readings do come from the thoughts.
On the twin pairs, our decoded thought shows the twin's values on 81\% of the pairs, against 25 to 45\% for the baselines, and it both shows them and is followed by the answer on 41\% and 42\%, against 6 to 16\%. When our answer follows the twin, the decoded thought shows it in over 90\% of the cases, against about 40\% for CODI and PCCoT, whose answers mostly follow the twin without the reading showing it. Coconut shows the opposite pattern, its thought showing the twin's values on 45\% of the pairs while its answer follows the twin on only 9\%.
The alignment of our models remains partial, since on about 40\% of the pairs the decoded thought shows a twin value but the answer is not the twin's, mostly because the twin's thought goes wrong after that value and the answer lands on neither.

\subsection{Refinable: Accuracy against Inference Budget}
\label{app:refinable}

\begin{figure}[t]
    \centering
    \includegraphics[width=\linewidth]{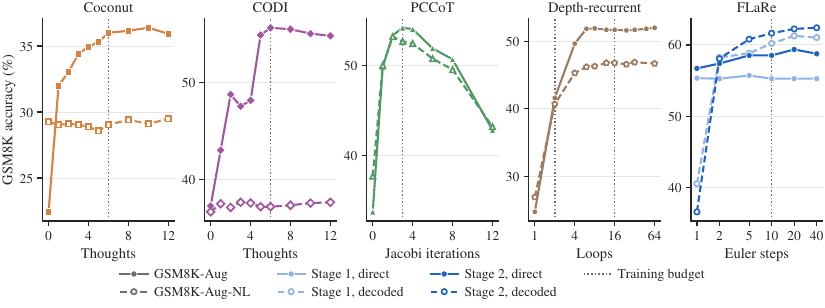}
    \caption{Accuracy against the latent computation spent at inference: thoughts for Coconut and CODI, Jacobi iterations for PCCoT, loops for the depth-recurrent model, and Euler steps for our flow model, on GSM8K-Aug and GSM8K-Aug-NL. The dotted lines mark the training budget.}
    \label{fig:budget}
\end{figure}

\textbf{Probe.}
We run each frozen model below and beyond its training budget: 0 to 12 thoughts for Coconut and CODI (trained with 6), 0 to 12 Jacobi iterations for PCCoT (trained with 3), and 1 to 40 Euler steps for ours (trained with 10 in stage 2).
The refinable column of \tabref{tab:properties} puts all methods on the compute scale of explicit CoT, 24 sequential passes, and reports the gain from about a tenth to half of it: 2 to 12 thoughts or iterations for the baselines, and 1 to 5 Euler steps for ours, that is 2 to 10 passes with the two guidance branches.
\figref{fig:budget} also shows each baseline trained on GSM8K-Aug-NL, which writes the same steps as sentences, and a depth-recurrent model~\citep{geiping2025scaling,mcleish2025retrofitted} that we train ourselves, looping the middle layers of the same base model 2 to 16 times.

\textbf{Results.}
Results are shown in \figref{fig:budget}. Coconut stays flat, CODI loses about a point at twice its budget, PCCoT drops from 54.1 at three iterations to 42.8 at twelve, and the depth-recurrent model is flat from six loops to 64.
On the decoded reading, our stage 1 model rises from 40.6 at one step to 58.2 at two and 61.3 at twenty, while its direct reading is 55.3 from the first step on. Stage 2, whose answer pass is trained on the 10-step rollout, gains on both readings from one to twenty steps, from 36.6 to 62.6 decoded and from 56.7 to 59.1 direct. Neither model loses more than a point at forty steps.

\subsection{Efficient: Latency Measurement}
\label{app:efficient}
\label{app:latency}

\begin{figure}[t]
\begin{minipage}[c]{0.53\linewidth}
\captionof{table}{Cost per GSM8K question: accuracy (\%), sequential passes before the first answer token, generated tokens and median latency (ms) over 100 questions. \ours: stage 2, direct at $S$ Euler steps, decoded at 20.}
\label{tab:cost}
\vspace{4pt}
\centering
\small\setlength{\tabcolsep}{1.5pt}
\begin{tabular}{@{}lcccc@{}}
\toprule
\textbf{Method} & \textbf{Acc.} & \textbf{Passes} & \textbf{Tokens} & \textbf{Latency} \\
\midrule
No-CoT & 30.3 & 0 & 3.2 & 40 \\
Explicit CoT & 59.3 & 24 & 27.6 & 267 \\
Coconut, 6 thoughts & 36.0 & 6 & 3.2 & 104 \\
CODI, 6 thoughts & 55.6 & 6 & 6.2 & 136 \\
PCCoT, 3 iterations & 54.1 & 3 & 1.2 & 69 \\
\midrule
\ours, direct, $S=1$ & 56.7 & 2 & 2.2 & \textbf{47} \\
\ours, direct, $S=2$ & 57.4 & 4 & 2.2 & 68 \\
\ours, direct, $S=5$ & 58.5 & 10 & 2.3 & 133 \\
\ours, direct, $S=10$ & 58.5 & 20 & 2.3 & 242 \\
\ours, direct, $S=20$ & 59.1 & 40 & 2.2 & 461 \\
\ours, decoded, $S=20$ & \textbf{62.6} & 40 + 26 & 25.8 & 959 \\
\bottomrule
\end{tabular}
\end{minipage}\hfill
\begin{minipage}[c]{0.44\linewidth}
    \centering
    \includegraphics[width=\linewidth]{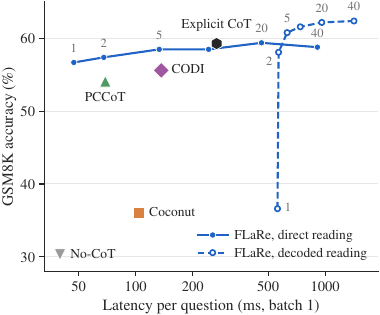}
    \caption{Accuracy against latency per question. Our curves vary the number of Euler steps, shown beside each point.}
    \label{fig:cost}
\end{minipage}
\end{figure}
    
\textbf{Probe.}
We time each method from the question to its last answer token, one question at a time on one RTX 3090 in bf16, and report the median over 100 test questions. Each baseline runs its reference inference code, and ours runs the stage 2 model with $S$ Euler steps of two forward passes each, followed by the direct reading or by the decoded reading, which adds greedy decoding by the frozen 3B decoder.

\textbf{Results.}
Results are shown in \tabref{tab:cost} and \figref{fig:cost}. One Euler step costs 22\,ms and one token of explicit CoT 9.7\,ms, so twenty steps cost as much as a CoT of about 45 tokens.
Our direct curve lies above and to the left of CODI and PCCoT from the first step, at 56.7 in 47\,ms, and flattens from five steps, where it is within a point of explicit CoT at half its latency. At two steps, it reaches 57.4 in 68\,ms, at half the latency of CODI and a quarter of that of explicit CoT, and at twenty steps it nearly matches explicit CoT at 1.7 times its latency.
The decoded reading serves only as a probe and is never used to answer at inference. Its curve is shifted right by half a second of decoding, more than the reasoning itself. It starts at 36.6 after a single step and rises to 60.8 at five and 62.6 at twenty, above every other point from five steps on. Since the decoder reads the thought alone, without the question, this rise shows that the extra steps refine the thought itself: after one step it is a rough estimate that often decodes into a wrong CoT, and each further step brings it closer to a correct one. The direct reading, in contrast, gains less than three points after its first step, since stage 1 trains the answer pass on the model's own one-step endpoints, so it can read the answer from a rough thought.

\bibliographystyle{plainnat}
\bibliography{references}

\end{document}